\documentclass{article}

    \PassOptionsToPackage{numbers, compress}{natbib}

\usepackage[eandd, final]{neurips_2026}

\usepackage[utf8]{inputenc} 
\usepackage[T1]{fontenc}    
\usepackage{hyperref}       
\usepackage{url}            
\usepackage{booktabs}       
\usepackage{amsfonts}       
\usepackage{nicefrac}       
\usepackage{microtype}      
\usepackage{xcolor}         

\usepackage{amsmath}
\usepackage{enumitem}

\usepackage[breakable]{tcolorbox}
\usepackage{listings}
\usepackage{float}
\usepackage[table]{xcolor}
\usepackage{stfloats}
\usepackage{multirow}
\usepackage{wrapfig}
\usepackage{amssymb}
\newcommand{\github}{\raisebox{-1.5pt}{\includegraphics[height=1.05em]{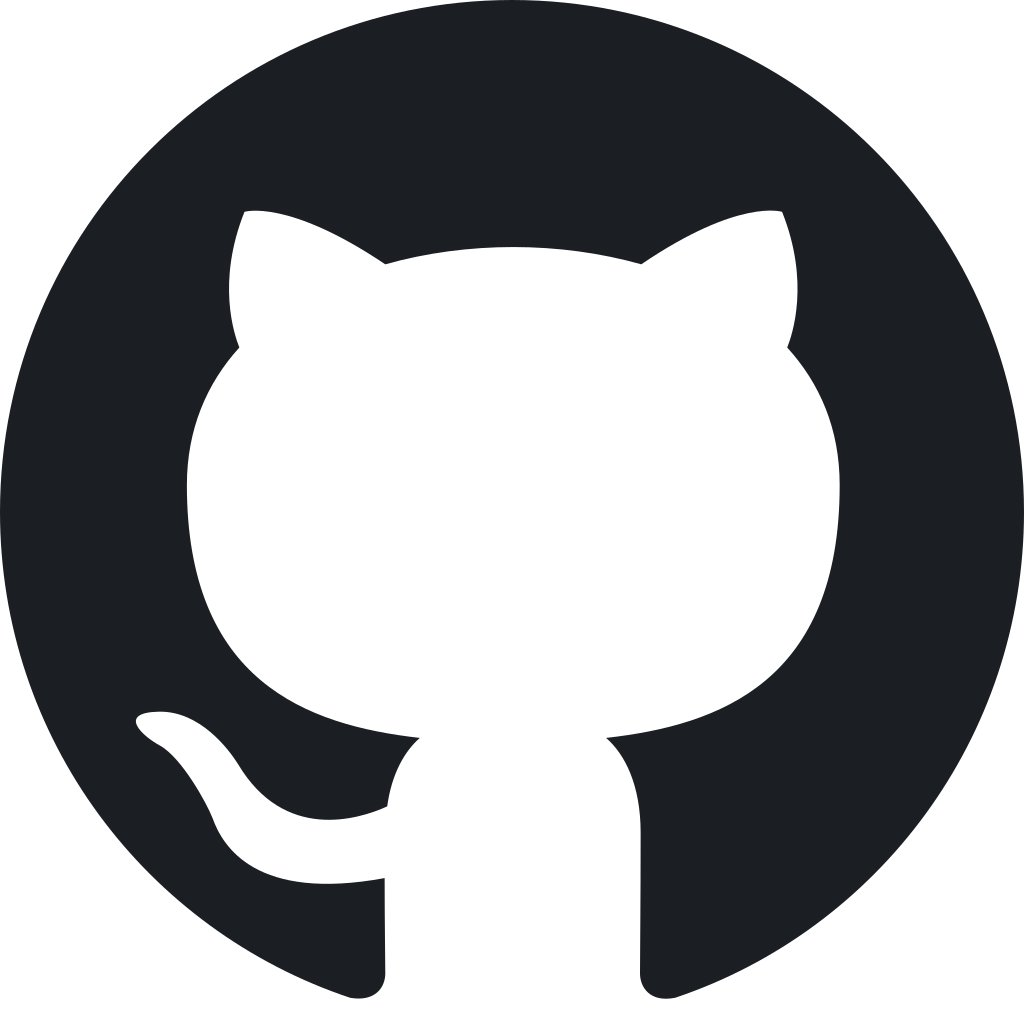}}}

\title{Measuring and Mitigating the Distributional Gap Between Real and Simulated User Behaviors}

\author{
    \textbf{Shuhaib Mehri$^{\spadesuit}$} \quad
    \textbf{Philippe Laban$^{\diamondsuit}$} \quad
    \textbf{Sumuk Shashidhar$^{\spadesuit}$} \quad
    \textbf{Marwa Abdulhai$^{\bigstar}$} \quad \\
    \textbf{Sergey Levine$^{\bigstar}$} \quad
    \textbf{Michel Galley$^{\diamondsuit}$} \quad
    \textbf{Dilek Hakkani-T\"ur$^{\spadesuit}$}
    \vspace{0.4em} \\
    \textnormal{$^{\spadesuit}$ University of Illinois Urbana-Champaign \quad $^{\diamondsuit}$} Microsoft Research \\
    \textnormal{$^{\bigstar}$ University of California, Berkeley}
    \vspace{0.4em} \\
    \texttt{mehri2@illinois.edu}
    \vspace{0.4em} \\
    \github \textbf{ Code}: \url{https://github.com/shuhaibm/UserBehavioralDivergence}
}

\begin{document}

\maketitle

\begin{abstract}
As user simulators are increasingly used for interactive training and evaluation of AI assistants, it is essential that they represent the diverse behaviors of real users. While existing works train user simulators to generate human-like responses, whether they capture the broad and heterogeneous distribution of real user behaviors remains an open question. In this work, we introduce a method to measure the distributional gap between real and simulated user behaviors, validated through a human study and ablations. Given a dataset of real and simulated conversations, our method extracts representations of user behavior from each conversation, quantizes them into discrete distributions via clustering, then computes divergence metrics. We provide the first systematic evaluation of 24 LLM-based user simulators on coding and writing tasks, and reveal a large distributional gap from real users that varies across model families, scales, and behavioral facets. Pairwise comparisons show that most simulators behave similarly, while a few stand apart. Combining behaviorally complementary simulators brings the resulting distribution closer to real users compared to either simulator on its own. Finally, a TF-IDF analysis of the clusters surfaces interpretable patterns of behaviors that simulators capture, miss, and hallucinate.
\end{abstract}
\begin{figure*}[h]
    \centering
    \vspace{-1em}
    \includegraphics[width=\textwidth]{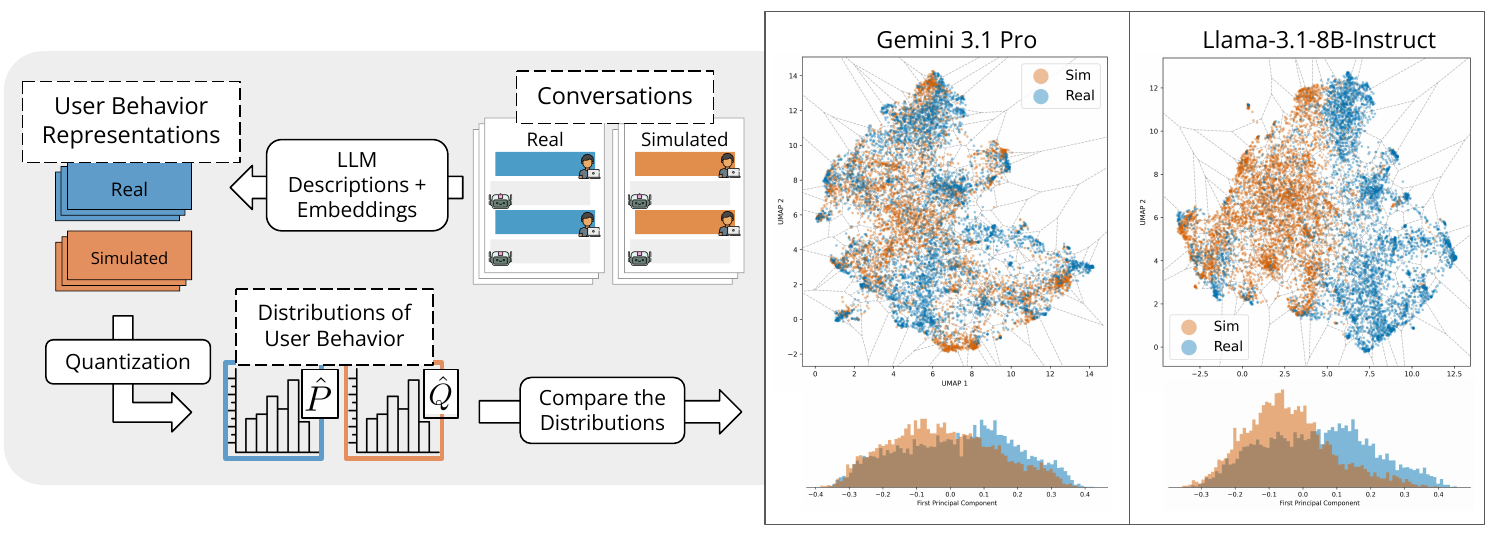}
    \vspace{-1.5em}
    \caption{
    Our method extracts representations of user behavior from real and simulated conversations, then quantizes them via $k$-means to get discrete behavioral distributions for real ($\hat{P}$) and simulated users ($\hat{Q}$). UMAP projections \citep{McInnes2018} and first principal component histograms illustrate the distributional gap for two simulators on the coding task: \texttt{Gemini 3.1 Pro} overlaps real users more closely compared to \texttt{Llama-3.1-8B-Instruct}.
    }
    \label{fig:intro}
    \vspace{-1em}
\end{figure*}

\section{Introduction}

The role of user simulators is to model the diverse behaviors of real users, providing a scalable means to train and evaluate AI systems on the interactions that they encounter in real-world settings \citep{levin2000stochastic, eckert1997user, pietquin2013survey, rieser2006cluster}. Recent works build user simulators by training Large Language Models (LLMs) to generate responses that resemble those of real users \citep{naous2026flipping, gandhi2026learning, wu2026humanlm}. Yet real users exhibit a broad, heterogeneous distribution of behaviors \citep{zhao2024wildchat, zheng2024lmsyschatm, liu2023one}. For example, some users tend to underspecify their requests, while others fully specify every constraint. The extent to which user simulators capture this distribution has not been measured.

A user simulator can fail to capture the distribution of real users in two ways: by demonstrating behaviors that real users rarely exhibit \textit{(low precision)}, or by failing to demonstrate behaviors that real users do exhibit \textit{(low recall)}, mirroring precision and recall metrics for generative models \citep{sajjadi2018assessing, kynkaanniemi2019improved, djolonga2020precision}. These failure modes bias training and evaluation, and yield assistants that struggle to generalize to the diverse behaviors of real users \citep{shi2019build}.

We introduce a method to compare the distributions of user behaviors in real and simulated conversations, illustrated in Figure~\ref{fig:intro}. For each conversation, we extract a representation of the user's behavior by using an LLM to generate a description along six behavioral facets (such as how they make requests or what dialog acts they perform) and then embedding it into a shared semantic space. Next, we quantize these representations into discrete distributions over user behavior modes by clustering \citep{sajjadi2018assessing, pillutla2021mauve}. Finally, we measure the gap between the real and simulated distributions using divergence-based metrics. We confirm that our method captures meaningful behavioral distributions and is robust to the choice of LLM, embedding model, clustering algorithm, and number of clusters through a human study and multiple validation studies.

We provide the first systematic evaluation of 24 LLM-based user simulators across coding and writing tasks, including 7 closed-source LLMs, 15 open-source LLMs, and 2 trained simulators. Our results reveal a large distributional gap from real users, and surface insights across model families, scales, and behavioral facets. Pairwise comparisons between simulators show that most behave similarly, while a few stand apart. Combining behaviorally complementary simulators brings the resulting distribution closer to real users than either simulator on its own. Lastly, a TF-IDF analysis of the clusters provides interpretable insights into behaviors that user simulators capture, miss, and hallucinate.

Our work advances toward user simulators that faithfully represent the diversity of real users with the following contributions: (1) We introduce a method for measuring the distributional gap between real and simulated user behaviors, validated through a human study and ablations; (2) Our analysis of 7 closed-source LLMs, 15 open-source LLMs and 2 trained simulators on coding and writing tasks reveals a large distributional gap from real users that varies across model families, scales, and behavioral facets; (3) We show that combining behaviorally complementary simulators brings the resulting distribution closer to real users than either simulator on its own; (4) We present a TF-IDF analysis of behavior clusters that surfaces interpretable insights into behaviors that simulators capture, miss, and hallucinate.
\section{Problem Formulation}

In this section, we formalize the problem of evaluating how well a user simulator captures the distribution of real user behaviors. Let $P$ denote the distribution of real user behaviors, and $Q$ denote the distribution of simulated user behaviors. Our objective is to measure the gap between $P$ and $Q$.

In practice, $P$ and $Q$ are not directly observable. Instead, we access them through datasets of conversations, where each conversation is a sequence $\mathcal{C}_n = (u_1, a_1, \ldots, u_n, a_n)$, with $u_i$ and $a_i$ denoting the user and assistant utterances at turn $i$. We use $\mathcal{D}_{\text{real}}$, a dataset of real user-assistant conversations, as samples from $P$. 

To sample from $Q$, we use an LLM to extract the user goal $\mathcal{G}$ (a description of the user's overall objective) from each conversation in $\mathcal{D}_{\text{real}}$ and provide it to the user simulator. The simulator generates utterances $u_i$ conditioned on $\mathcal{G}$ and the conversation history $\mathcal{C}_{i-1}$, and a fixed assistant generates utterances $a_i$ conditioned on the conversation history $\mathcal{C}_{i-1}$ and $u_i$. Conversations terminate upon reaching a maximum length or when the simulator emits a termination signal (e.g., \texttt{<|TERMINATE\_CONVERSATION|>}). The resulting conversations yield $\mathcal{D}_{\text{sim}}$, our samples from $Q$. Measuring the distributional gap between $P$ and $Q$ now reduces to comparing the user behaviors observed in $\mathcal{D}_{\text{real}}$ with those generated by the user simulator in $\mathcal{D}_{\text{sim}}$.
\section{Method}
\label{sec:method}

We present a three-stage method for measuring the distributional gap between the user behaviors in $\mathcal{D}_{\text{real}}$ and $\mathcal{D}_{\text{sim}}$, illustrated in Figure~\ref{fig:intro}: (1) Generating User Behavior Representations, (2) Quantizing into Behavioral Distributions, and (3) Measuring the Distributional Gap.

\textbf{Step 1: Generating User Behavior Representations}

The first step is to extract a representation of the user's behavior from each conversation in $\mathcal{D}_{\text{real}}$ and $\mathcal{D}_{\text{sim}}$. To do so, we first prompt an LLM to generate a description of the user's behavior along six facets, grounded in established frameworks for analyzing user behavior \citep{taylor1967question, allen1997draft, young2010hidden, henderson-etal-2014-second, pietquin2013survey}. The first four facets (Requests, Responses, Context, Communication Style) capture conversation-level behaviors, while the last two (DAMSL and SGD Dialog Acts) capture utterance-level behaviors through dialog act annotation frameworks:

\begin{itemize}[leftmargin=12pt,topsep=0pt]
    \item \textbf{Requests:} What types of requests users make to the assistant and how. We consider how explicit and specified each request is \citep{taylor1967question, piantadosi2012communicative, andukuri2024stargate, zamfirescu2023johnny}, how the user goal is decomposed across turns \citep{grosz1986attention, henderson-etal-2014-second}, and whether requests serve the primary goal or secondary functions \citep{mehri2025goal}.
    \item \textbf{Responses:} How users respond to the assistant. We consider their engagement levels \citep{zhao2024wildchat, zheng2024lmsyschatm}, how they evaluate assistant outputs \citep{shelby2025taxonomy}, the type of feedback they provide \citep{shi2025wildfeedback}, and whether they introduce new constraints or preferences across turns \citep{mehri2026learning, li2025eliciting}.
    \item \textbf{Context:} How users provide background information. We consider the type of context they provide (e.g., domain knowledge, prior attempts, thought processes, personal background) \citep{taylor1967question}, how directly it relates to the user goal \citep{mehri2025goal}, whether context is front-loaded or revealed gradually across turns \citep{clark1991grounding}, and whether it is volunteered proactively or elicited by the assistant \citep{horvitz1999principles, wu2025collabllm}.
    \item \textbf{Communication Style:} How users communicate stylistically. We consider their register and emotional tone, verbosity, message formatting (e.g., bullet points, markdown, prose), and social conventions such as politeness and pleasantries \citep{pietquin2013survey, zhou2026mind, naous2026flipping, hoang2026psibenchclinicallygroundedinterpretable, zhou2026mind}.
    \item  \textbf{DAMSL Dialog Acts:} Per-utterance analysis using the Dialog Act Markup in Several Layers (DAMSL) framework. It characterizes utterances across three aspects: information level (the semantic content of the utterance), forward-looking function (its effect on subsequent dialog), and backward-looking function (relation to prior discourse) \citep{allen1997draft}.
    \item  \textbf{SGD Dialog Acts:} Per-utterance classification into one or more discrete dialog act labels adapted from the Schema-Guided Dialogue (SGD) dataset, such as inform, request, and affirm \citep{rastogi2020towards, young2010hidden}.
\end{itemize}

These behavioral facets abstract away irrelevant features, enabling our comparisons to focus on behavioral patterns rather than surface-level signals such as lexical similarity. The full prompts and criteria are provided in Appendix~\ref{appendix:user_behavior_representations}.

For each conversation, we concatenate descriptions across all six facets into a single textual representation of user behavior. Then, we embed each representation with a text embedding model, mapping behaviors from $\mathcal{D}_{\text{real}}$ and $\mathcal{D}_{\text{sim}}$ into a shared semantic space. As \citet{pimentel2023on} show, embeddings capture discourse- and coherence-level features while ignoring surface-level ones, making them well-suited for representing user behaviors.

\textbf{Step 2: Quantizing into Behavioral Distributions}

The user behavior representations for $\mathcal{D}_{\text{real}}$ and $\mathcal{D}_{\text{sim}}$ are finite sets of continuous, high-dimensional vectors. This makes estimating divergences unreliable. Following the precision and recall framework of \citet{sajjadi2018assessing} and the MAUVE methodology of \citet{pillutla2021mauve}, we employ a quantization step to map the representations into low-dimensional discrete distributions of behaviors. We apply $k$-means clustering to the set of user behavior representations from $\mathcal{D}_{\text{real}}$ and $\mathcal{D}_{\text{sim}}$. Each of the $k$ clusters groups representations with similar behavioral patterns, and thus represents a particular mode of user behavior. We obtain probability distributions $\hat{P}$ and $\hat{Q}$ over $c \in \{1, \ldots, k \}$, where $\hat{P}(c)$ and $\hat{Q}(c)$ denote the fraction of representations from $\mathcal{D}_{\text{real}}$ and $\mathcal{D}_{\text{sim}}$ in cluster $c$. These distributions capture how frequently different user behavior modes occur in real and simulated conversations, and serve as estimates of $P$ and $Q$.

\textbf{Step 3: Measuring the Distributional Gap}

Given $\hat{P}$ and $\hat{Q}$, we measure the gap between the behavioral distributions to understand how well the user simulator represents real users. The simulator can diverge from real users in two ways: by demonstrating behaviors that real users rarely exhibit \textit{(low precision)} or by failing to demonstrate behaviors that real users do exhibit \textit{(low recall)}, mirroring precision and recall metrics in generative modeling \citep{sajjadi2018assessing, kynkaanniemi2019improved, djolonga2020precision}. Formally, low precision arises when $\hat{Q}$ assigns high probability to behaviors that are rare under $\hat{P}$, while low recall arises when $\hat{Q}$ assigns low probability to behaviors that are common under $\hat{P}$. This formulation captures the intuitions of bias and variance in simulator behavior: reductions in behavioral variance manifest as low recall, and excess variance into behaviors real users do not exhibit manifests as low precision. We report the following metrics:

\begin{itemize}[leftmargin=12pt,topsep=0pt]
    \item \textbf{Forward KL Divergence:} Defined as $\mathrm{KL}(\hat{P}\,\|\,\hat{Q}) = \sum_c \hat{P}(c) \log\frac{\hat{P}(c)}{\hat{Q}(c)}$. Higher values indicate low recall, meaning the simulator fails to demonstrate behaviors that real users exhibit.
    \item \textbf{Backward KL Divergence:} Defined as $\mathrm{KL}(\hat{Q}\,\|\,\hat{P}) = \sum_c \hat{Q}(c) \log\frac{\hat{Q}(c)}{\hat{P}(c)}$. Higher values indicate low precision, meaning the simulator demonstrates behaviors that real users do not exhibit.
    \item \textbf{Jensen--Shannon Divergence:} A symmetric divergence defined as $\mathrm{JS}(\hat{P}, \hat{Q}) = \tfrac{1}{2}\mathrm{KL}(\hat{P}\,\|\,\hat{M}) + \tfrac{1}{2}\mathrm{KL}(\hat{Q}\,\|\,\hat{M})$, where $\hat{M} = \tfrac{1}{2}(\hat{P} + \hat{Q})$. JS divergence captures both low precision and low recall.
\end{itemize}
\section{Experimental Setup}
\label{sec:experimental_setup}

We evaluate 24 user simulators: 7 closed-source LLMs, 15 open-source LLMs, and 2 trained simulators. For each user simulator, we measure the distributional gap with real users by instantiating $\mathcal{D}_{\text{real}}$ from real user-assistant conversations and generating the corresponding $\mathcal{D}_{\text{sim}}$.

$\mathcal{D}_{\text{real}}$ is constructed from WildChat \citep{zhao2024wildchat}, a dataset of one million real-world user-assistant interactions. We use the postprocessed subset from \citet{naous2026flipping}, which filters to English-only conversations, deduplicates, and provides GPT-4o generated user goals $\mathcal{G}$ for each conversation. Each $\mathcal{G}$ is classified by task with \texttt{Qwen3.5-122B-A10B-FP8} \citep{qwen35blog} (see Appendix~\ref{appendix:user_goal_classification}). We focus on coding, where users aim to produce functional code, and writing, where users aim to produce a written artifact. We sample 5,000 conversations to get $\mathcal{D}_{\text{real}}^{\text{coding}}$ and $\mathcal{D}_{\text{real}}^{\text{writing}}$. To generate $\mathcal{D}_{\text{sim}}^{\text{coding}}$ and $\mathcal{D}_{\text{sim}}^{\text{writing}}$, we provide each user goal $\mathcal{G}$ from $\mathcal{D}_{\text{real}}^{\text{coding}}$ and $\mathcal{D}_{\text{real}}^{\text{writing}}$ to the simulator and generate the corresponding simulated conversation. We use \texttt{Qwen3.5-122B-A10B} as the assistant. All prompts are provided in Appendix~\ref{appendix:conversation_generation_prompts}.

To establish the lower bound, we measure the divergence between two behavioral distributions of real users. For each conversation in $\mathcal{D}_{\text{real}}^{\text{coding}}$ and $\mathcal{D}_{\text{real}}^{\text{writing}}$, we sample its match from a held-out WildChat subset based on user-goal embedding similarity, giving us a similar conversation with a different user.

\paragraph{Method Configuration.} We generate behavior descriptions with \texttt{Qwen3.5-122B-A10B-FP8} and embed them using \texttt{Qwen3-Embedding-8B} \citep{qwen3embedding} (truncated to 1024 dimensions). For quantization, we follow a similar implementation to \citet{pillutla2021mauve}: we concatenate the real and simulated sets of embeddings, $\ell_2$-normalize each embedding, and reduce dimensionality via PCA to 90\% explained variance. Then, we run $k$-means with $k = 500$ for up to $500$ iterations across $5$ restarts, keeping the restart with the best objective. We apply Laplace smoothing with $\alpha = 1/k$ to all KL-based metrics.
\section{Results}
\label{sec:results}

\definecolor{okabe_orange}{HTML}{E69F00}
\definecolor{okabe_blue}{HTML}{56B4E9}
\definecolor{okabe_pink}{HTML}{CC79A7}
\providecommand{\ccol}[1]{\cellcolor[HTML]{#1}}

\begin{table*}
\tiny
  \caption{
    The distributional gap between real and simulated user behaviors across coding and writing tasks. We compare the user behavioral distribution from $\mathcal{D}_{\text{real}}$, a subset of 5,000 conversations from WildChat \citep{zhao2024wildchat} with those from $\mathcal{D}_{\text{sim}}$, the corresponding simulated conversations. We report forward KL ($\mathrm{KL_{fwd}}$), backward KL ($\mathrm{KL_{bwd}}$), and JS divergence ($\mathrm{JS}$). $\uparrow$ indicates higher is better and $\downarrow$ indicates lower is better. Darker blue indicates closer behavioral distribution to real users.}
    \vspace{1em}
  \label{tab:main_results}
  \centering
  \resizebox{\textwidth}{!}{%
  \begin{tabular}{l ccc ccc}
    \toprule
    & \multicolumn{3}{c}{\textbf{Coding}} & \multicolumn{3}{c}{\textbf{Writing}} \\
    \cmidrule(lr){2-4} \cmidrule(lr){5-7}
    \textbf{User Simulator}

      & $\mathrm{KL_{\mathrm{fwd}}}$\,$\downarrow$
      & $\mathrm{KL_{\mathrm{bwd}}}$\,$\downarrow$
      & $\mathrm{JS}$\,$\downarrow$
      & $\mathrm{KL_{\mathrm{fwd}}}$\,$\downarrow$
      & $\mathrm{KL_{\mathrm{bwd}}}$\,$\downarrow$
      & $\mathrm{JS}$\,$\downarrow$ \\
    \specialrule{0.9pt}{2pt}{2pt}
    \rowcolor{okabe_blue} \textbf{\textit{Real Users}}
            & .025 & .025 & .028
            & .016 & .017 & .020 \\
    \specialrule{0.9pt}{2pt}{2pt}
    \rowcolor{okabe_pink!20}  \multicolumn{7}{c}{\textit{Closed-Source}} \\
    \midrule
    \texttt{GPT-5.4}
            & \ccol{79C4EE} .261 & \ccol{88CAF0} .265 & \ccol{88CAF0} .248
            & \ccol{79C4EE} .482 & \ccol{88CAF0} .489 & \ccol{88CAF0} .479 \\
    \texttt{GPT-5.4 mini}
            & \ccol{BAE0F6} .363 & \ccol{C5E5F8} .388 & \ccol{BAE0F6} .364
            & \ccol{93CFF1} .513 & \ccol{A5D7F3} .533 & \ccol{93CFF1} .520 \\
    \texttt{GPT-5.4 nano}
            & \ccol{A5D7F3} .353 & \ccol{ACDAF4} .364 & \ccol{B3DDF5} .353
            & \ccol{A5D7F3} .520 & \ccol{9DD3F2} .531 & \ccol{A5D7F3} .532 \\
    \texttt{Claude Haiku 4.5}
            & \ccol{E3F3FB} .438 & \ccol{DEF1FB} .430 & \ccol{DEF1FB} .427
            & \ccol{F7FCFE} .605 & \ccol{ECF6FC} .608 & \ccol{F0F8FD} .613 \\
    \texttt{Gemini 3.1 Pro}
            & \ccol{56B4E9} .260 & \ccol{56B4E9} .256 & \ccol{56B4E9} .246
            & \ccol{79C4EE} .482 & \ccol{79C4EE} .471 & \ccol{79C4EE} .466 \\
    \texttt{Gemini 3 Flash}
            & \ccol{ACDAF4} .355 & \ccol{9DD3F2} .328 & \ccol{9DD3F2} .334
            & \ccol{BAE0F6} .539 & \ccol{ACDAF4} .545 & \ccol{ACDAF4} .533 \\
    \texttt{Gemini 3.1 Flash-Lite}
            & \ccol{D5ECFA} .432 & \ccol{D0EAF9} .406 & \ccol{D5ECFA} .421
            & \ccol{DAEFFA} .583 & \ccol{D5ECFA} .586 & \ccol{CBE8F8} .582 \\
    \midrule
    \rowcolor{okabe_blue!20} \multicolumn{7}{c}{\textit{Open-Source}} \\
    \midrule
    \texttt{Qwen3.5-122B-A10B}
            & \ccol{C0E3F7} .394 & \ccol{BAE0F6} .378 & \ccol{C5E5F8} .387
            & \ccol{D5ECFA} .581 & \ccol{C0E3F7} .575 & \ccol{C5E5F8} .579 \\
    \texttt{Qwen3.5-35B-A3B}
            & \ccol{F4FAFE} .480 & \ccol{F4FAFE} .467 & \ccol{F7FCFE} .478
            & \ccol{F4FAFE} .602 & \ccol{F0F8FD} .609 & \ccol{F4FAFE} .615 \\
    \texttt{Qwen3.5-27B}
            & \ccol{DEF1FB} .437 & \ccol{D5ECFA} .421 & \ccol{E3F3FB} .433
            & \ccol{C5E5F8} .570 & \ccol{CBE8F8} .580 & \ccol{D5ECFA} .584 \\
    \texttt{Qwen3.5-9B}
            & \ccol{ECF6FC} .465 & \ccol{F0F8FD} .457 & \ccol{ECF6FC} .454
            & \ccol{CBE8F8} .571 & \ccol{C0E3F7} .575 & \ccol{CBE8F8} .582 \\
    \texttt{Qwen3.5-4B}
            & \ccol{F7FCFE} .482 & \ccol{F7FCFE} .471 & \ccol{F4FAFE} .472
            & \ccol{E7F4FC} .597 & \ccol{F0F8FD} .609 & \ccol{ECF6FC} .609 \\
    \texttt{Qwen3.5-2B}
            & \ccol{FFFFFF} .584 & \ccol{FFFFFF} .591 & \ccol{FFFFFF} .608
            & \ccol{FFFFFF} .630 & \ccol{FFFFFF} .642 & \ccol{FFFFFF} .660 \\
    \texttt{Qwen3.5 .8B}
            & \ccol{FBFDFF} .547 & \ccol{FBFDFF} .568 & \ccol{FBFDFF} .579
            & \ccol{FBFDFF} .629 & \ccol{FBFDFF} .620 & \ccol{FBFDFF} .648 \\
    \texttt{Llama-3.3-70B-Instruct}
            & \ccol{F0F8FD} .471 & \ccol{ECF6FC} .456 & \ccol{F0F8FD} .469
            & \ccol{F0F8FD} .600 & \ccol{E7F4FC} .602 & \ccol{F4FAFE} .615 \\
    \texttt{Llama-3.1-8B-Instruct}
            & \ccol{DAEFFA} .435 & \ccol{DAEFFA} .427 & \ccol{DAEFFA} .423
            & \ccol{ECF6FC} .598 & \ccol{F7FCFE} .610 & \ccol{E7F4FC} .608 \\
    \texttt{gpt-oss-120b}
            & \ccol{B3DDF5} .358 & \ccol{A5D7F3} .353 & \ccol{A5D7F3} .343
            & \ccol{ACDAF4} .532 & \ccol{BAE0F6} .549 & \ccol{B3DDF5} .537 \\
    \texttt{gpt-oss-20b}
            & \ccol{CBE8F8} .411 & \ccol{E3F3FB} .435 & \ccol{D0EAF9} .409
            & \ccol{C0E3F7} .569 & \ccol{E3F3FB} .593 & \ccol{C0E3F7} .575 \\
    \texttt{gemma-4-31B-it}
            & \ccol{93CFF1} .311 & \ccol{93CFF1} .296 & \ccol{93CFF1} .297
            & \ccol{B3DDF5} .537 & \ccol{B3DDF5} .547 & \ccol{BAE0F6} .545 \\
    \texttt{gemma-4-26B-A4B-it}
            & \ccol{C5E5F8} .397 & \ccol{B3DDF5} .373 & \ccol{C0E3F7} .379
            & \ccol{D0EAF9} .577 & \ccol{D0EAF9} .585 & \ccol{DAEFFA} .587 \\
    \texttt{gemma-4-E4B-it}
            & \ccol{D0EAF9} .422 & \ccol{C0E3F7} .383 & \ccol{CBE8F8} .402
            & \ccol{E3F3FB} .589 & \ccol{DEF1FB} .591 & \ccol{DEF1FB} .597 \\
    \texttt{gemma-4-E2B-it}
            & \ccol{E7F4FC} .446 & \ccol{E3F3FB} .435 & \ccol{E7F4FC} .435
            & \ccol{DEF1FB} .586 & \ccol{DAEFFA} .589 & \ccol{E3F3FB} .599 \\
    \midrule
    \rowcolor{okabe_orange!20}\multicolumn{7}{c}{\textit{Trained Simulators}} \\
    \midrule
    \texttt{UserLM-8b}
        & \ccol{9DD3F2} .328 & \ccol{C5E5F8} .388 & \ccol{ACDAF4} .349
            & \ccol{56B4E9} .392 & \ccol{56B4E9} .424 & \ccol{56B4E9} .411 \\
    \texttt{humanlm-opinion}
            & \ccol{88CAF0} .268 & \ccol{79C4EE} .257 & \ccol{79C4EE} .247
            & \ccol{9DD3F2} .514 & \ccol{93CFF1} .519 & \ccol{9DD3F2} .531 \\
    \bottomrule
  \end{tabular}}
\vspace{-3em}
\end{table*}

Our main results in Table~\ref{tab:main_results} reveal a large distributional gap across all simulators, which tends to be smaller on coding compared to writing. Closed-source models generally outperform open-source models, with \texttt{GPT-5.4} and \texttt{Gemini 3.1 Pro} achieving the smallest distributional gap to real users. However, the best open-source models, \texttt{gemma-4-31B-it} and \texttt{gpt-oss-120b}, are competitive and outperform several closed-source models.

The trained simulators are particularly notable, with \texttt{humanlm-opinion} \citep{wu2026humanlm} and \texttt{UserLM-8b} \citep{naous2026flipping} achieving results on par with the best closed-source models despite being 8B parameter models. However, we note that \texttt{UserLM-8b} was trained on WildChat \citep{zhao2024wildchat}, which overlaps with our evaluation data, so its results should be interpreted with caution. Meanwhile, a preliminary experiment on writing shows that \texttt{Qwen3.5-4B-Base} achieves lower divergence than its post-trained counterpart \texttt{Qwen3.5-4B} across all three metrics (Appendix~\ref{appendix:base_vs_posttrained}). Further understanding the dynamics of how post-training affects the ability of LLMs to simulate diverse user behaviors is a promising direction for future work.

Scale alone is insufficient to close the gap with real users. Only the Gemma and gpt-oss families show consistent improvement with scale. For Qwen3.5, the largest model (122B) achieves the highest performance, but the trend among the smaller models is inconsistent. In the Llama family, the smaller \texttt{Llama-3.1-8B-Instruct} outperforms \texttt{Llama-3.3-70B-Instruct}. These results indicate that other factors, such as training data or model family, matter alongside scale.

Results for each individual facet (Appendix~\ref{appendix:per_dimension_results}) provide further insights into where behavioral gaps originate. Simulators approximate the requests and context facets relatively well, but diverge on the communication style, DAMSL and SGD dialog acts. On these harder facets, the trained simulators outperform the other models by a wider margin, indicating that finetuning captures the behaviors that general purpose LLMs miss. These results also reveal fine-grained error patterns. For instance, forward KL tends to be greater than backward KL on SGD dialog acts for coding, indicating that simulators miss the diversity of dialog acts that real users exhibit. 

\section{Do Embeddings and Clusters Effectively Capture Distributions of User Behaviors?}
\label{sec:method_validation}

\subsection{"Odd-One-Out" Human Study}
\label{sec:method_validation_human_study}
For each conversation, our method generates user behavior representations and then quantizes them into discrete distributions using $k$-means. Conversations with similar user behavior are grouped in the same cluster. We validate this with an "odd-one-out" task \citep{NIPS2009_f92586a2}: annotators are shown three behavior descriptions, two from the same cluster and one from a different cluster, then asked to identify which does not belong. We randomly sample 25 triplets of user behavior descriptions and present them to 15 annotators. Participant details and instructions are in Appendix \ref{appendix:humman_study}. Annotators correctly identified the odd-one-out 86.7\% of the time on average, with high inter-annotator agreement (Fleiss' $\kappa = 0.74$). This confirms that clusters capture meaningful behavioral similarity. This study was declared exempt by our Institutional Review Board (IRB).

\subsection{Ablations}
\label{sec:method_validation_ablations}

\begin{wrapfigure}{h}{0.35\textwidth}
    \centering
    \vspace{-1em}
    \includegraphics[width=0.35\textwidth]{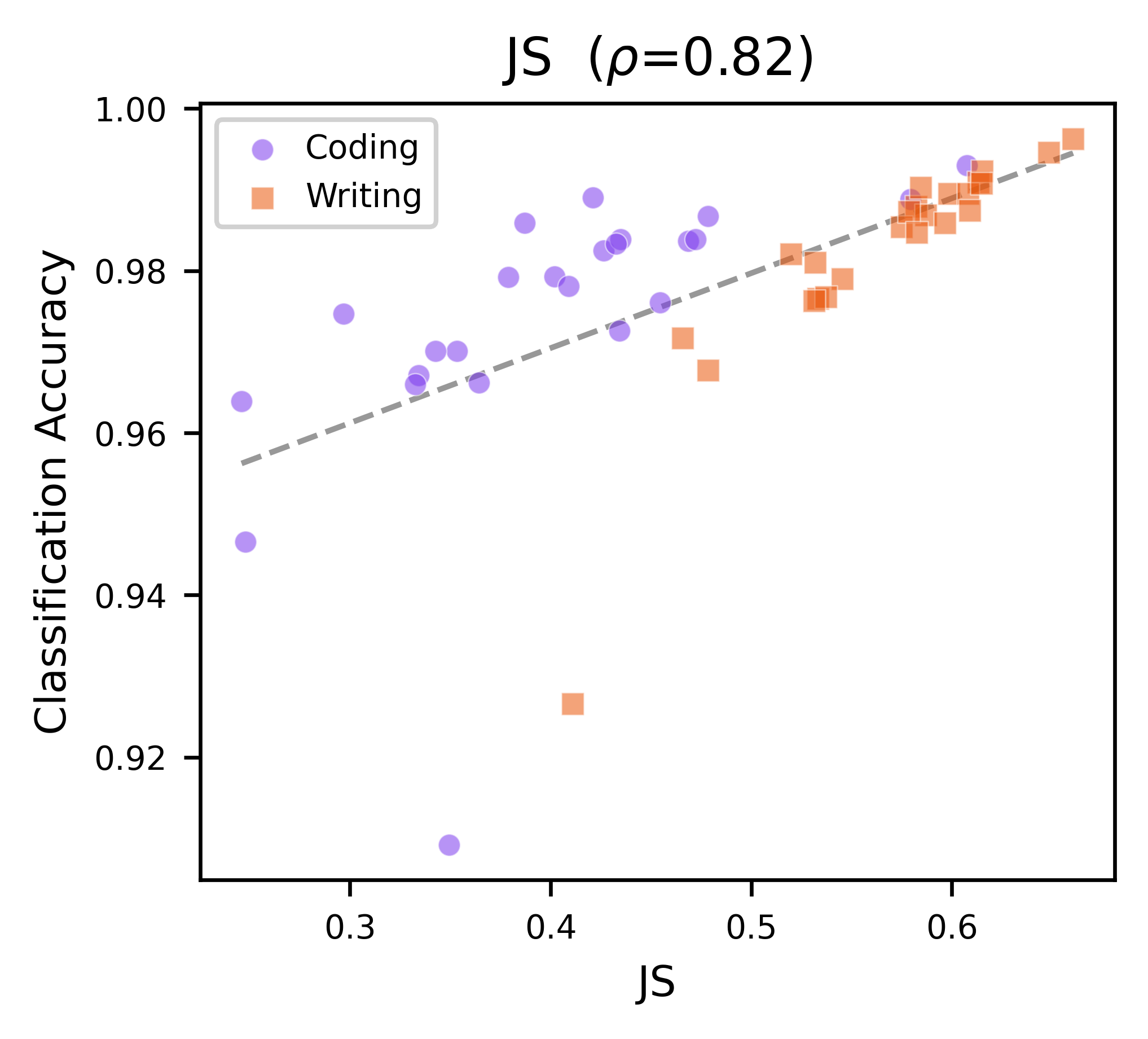}
    \vspace{-2em}
    \caption{Scatter plot of classification accuracy and Jensen--Shannon Divergence ($\mathrm{JS}$) for all simulators across coding and writing tasks. Each point represents one simulator on one task.}
    \vspace{-2em}
    \label{fig:classification_correlation_js}
\end{wrapfigure}

Our method generates user behavior representations before quantizing via $k$-means clustering to create discrete distributions. Each stage introduces a design choice that could influence the measured distributional gap. We conduct ablations to verify that our method is robust across different choices.

\paragraph{User Behavior Descriptions.} We use an LLM to generate user behavior descriptions for each conversation to isolate user behavioral patterns and abstract away irrelevant surface-level features from the raw conversation history. To validate that this step is necessary, we compare against two simpler representations: the raw conversations and user utterances only. Table~\ref{tab:representation_ablation} (Appendix \ref{appendix:ablation_results}) reports the full results for all simulators across both tasks. These simpler representations produce low divergences across all simulators, which would suggest that simulators closely match real users. However, this similarity is an artifact of surface-level features such as lexical and semantic overlap, rather than reflecting behavioral patterns. Our behavioral descriptions, by contrast, reveal much larger and more meaningful gaps, confirming that this step is necessary.

\begin{table}[t]
  \caption{
    Simulator rankings are robust across embedding models, clustering algorithms, and number of clusters $k$. We report Spearman rank correlation ($\rho$) between rankings under each design choice. Metric values are z-score normalized within each dataset before aggregation. Higher $\rho$ indicates greater robustness.}
  \label{tab:ablation}
  \centering
  \footnotesize
  \begin{tabular}{llccc}
    \toprule
    & & \multicolumn{3}{c}{\textbf{Spearman Rank Correlation ($\rho$)}} \\
    \cmidrule(lr){3-5}
    \textbf{Ablation} &\textbf{Pair} & $\mathrm{KL_{fwd}}$ & $\mathrm{KL_{bwd}}$ & $\mathrm{JS}$ \\
    \midrule
    \multirow{3}{*}{\textit{Embedding}}
        & \texttt{Qwen3-Embedding-8B} / \texttt{BGE-small-en-v1.5}
            & .86 & .84 & .87 \\
        & \texttt{Qwen3-Embedding-8B} / \texttt{e5-large-v2}
            & .87 & .80 & .90 \\
        & \texttt{BGE-small-en-v1.5} / \texttt{e5-large-v2}
            & .94 & .89 & .97 \\
    \midrule
    \multirow{3}{*}{\textit{Clustering Algorithm}}
        & \texttt{K-Means} / \texttt{Gaussian Mixture Models}
            & .96 & .95 & .97 \\
        & \texttt{K-Means} / \texttt{Agglomerative}
            & .98 & .96 & .98 \\
        & \texttt{Gaussian Mixture Models} / \texttt{Agglomerative}
            & .97 & .97 & .98 \\

    \midrule
    \multirow{5}{*}{\textit{Number of Clusters}}
        & $k=50$ / $k=500$  & .97 & .94 & .98 \\
        & $k=100$ / $k=500$  & .97 & .94 & .98 \\
        & $k=250$ / $k=500$  & .97 & .95 & .99 \\
        & $k=1000$ / $k=500$  & .97 & .95 & .99 \\
        & $k=2000$ / $k=500$  & .96 & .96 & 1.00 \\
            
    \bottomrule
  \end{tabular}
\end{table}

\paragraph{Embedding Models.} User behavior descriptions are embedded using \texttt{Qwen3-Embedding-8B}. To verify that our results are robust across embedding models, we repeat our experiments using two other embedding models: \texttt{e5-large-v2} \citep{wang2022text} and \texttt{BGE-small-en-v1.5} \citep{bge_embedding}. We apply z-score normalization to metric values within each dataset and then compute the Spearman rank correlation between simulator rankings produced by each pair of embedding models. Table~\ref{tab:ablation} report the results, and Figure~\ref{fig:ablations} (Appendix~\ref{appendix:ablation_results}) provide scatter plot visualizations. Simulator rankings are largely preserved across different embedding models.

\paragraph{Clustering.} We obtain discrete distributions of user behavior by clustering our embeddings using $k$-means with $k=500$. To verify that our method is robust to the clustering algorithm, we repeat our experiments using Gaussian Mixture Models and Agglomerative Clustering. Similar to the embedding model ablation, we compute the Spearman rank correlations between simulator rankings produced by each pair of algorithms. Simulator rankings are nearly identical across all pairs ($\rho \geq 0.95$), and scatter plot visualizations are provided in Figure~\ref{fig:ablations} (Appendix~\ref{appendix:ablation_results}). We further vary the number of clusters $k \in \{50, 100, 250, 1000, 2000\}$ and compare the resulting simulator rankings to those obtained with $k=500$, and find that rankings remain highly stable across all values of $k$ ($\rho \geq 0.94$). Table~\ref{tab:ablation} reports all correlation results.

\subsection{Linear Classification}

We validate that our user behavior representation embeddings encode meaningful differences between real and simulated users by training a linear classifier. For each simulator, we construct a dataset of 5,000 real and 5,000 simulated embeddings, and then train an L2-regularized logistic regression classifier to classify embeddings as real or simulated. We average results over 5 stratified random 80/20 train-test splits. The classification accuracies across simulators are consistently high, ranging from 90.92\% to 99.63\%, and also correlate strongly with our distributional metrics (Figure~\ref{fig:classification_correlation_js}). This confirms that the embeddings encode meaningful differences between real and simulated users. The full accuracy scores and correlations for all metrics are presented in Appendix~\ref{appendix:linear_classification}.
\section{Discussion}
\label{sec:discussion}

\subsection{How Do User Behavior Distributions Compare Among Simulators?}
\label{sec:cross_simulators}

\begin{figure*}[t]
    \centering
    \includegraphics[width=0.9\textwidth]{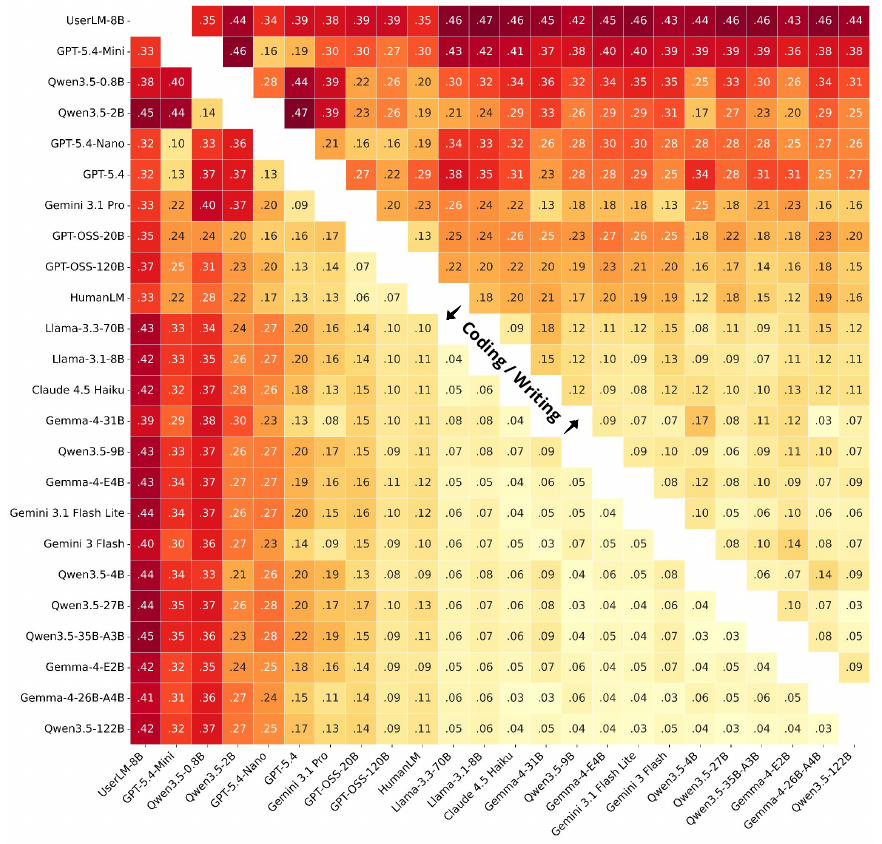}
    \caption{Pairwise Jensen--Shannon divergence between the behavior distributions of 24 user simulators across coding (lower triangle) and writing tasks (upper triangle), ordered by descending mean pairwise JS divergence. Many simulators demonstrate similar behavioral distributions (lighter cells) while a few stand out (darker cells).}
    \label{fig:cross_simulator}
    \vspace{-1em}
\end{figure*}

Our main results show that the distributional gap to real users varies across simulators. This raises a natural question: \textit{how do the behavioral distributions of user simulators compare to each other?} To investigate this, we pool the user behavior representations across all 24 simulators, quantize them into distributions, and compute the $\mathrm{JS}$ divergence between all pairs. Figure~\ref{fig:cross_simulator} reports the results.

Many simulators behave similarly to each other: 30\% of the 552 pairwise comparisons have $\mathrm{JS} < 0.10$. Pairs from the same family tend to be the most similar, achieving the lowest scores in the dataset. The simulators with the lowest mean pairwise $\mathrm{JS}$ scores are \texttt{Qwen3.5-122B-A10B} ($0.14$), \texttt{gemma-4-26B-A4B-it} ($0.15$), \texttt{gemma-4-E2B-it} ($0.15$), and \texttt{Qwen3.5-35B-A3B} ($0.15$). A few simulators stand out and demonstrate behaviors that diverge from the rest. These include \texttt{UserLM-8b}, which has the highest mean pairwise $\mathrm{JS}$ score of $0.41$, followed by \texttt{GPT-5.4 mini} with a score of $0.33$. Pairwise $\mathrm{JS}$ scores are consistently higher for writing than coding, which aligns with our main results, where the divergences from real users were also higher for writing.

\subsection{Can we Combine User Simulators to Reduce the Gap With Real Users?}

\definecolor{darkgreen}{rgb}{0.0, 0.5, 0.0}
\definecolor{darkred}{rgb}{0.5, 0.0, 0.0}
\definecolor{okabe_teal}{HTML}{009E73}
\providecommand{\ccol}[1]{\cellcolor[HTML]{#1}}

\begin{table*}
  \caption{The distributional gap between real and simulated user behaviors when combining pairs of simulators. When generating conversations, we randomly sample which LLM serves as the user simulator. We report forward KL divergence ($\mathrm{KL_{fwd}}$), backward KL divergence ($\mathrm{KL_{bwd}}$), and Jensen--Shannon divergence ($\mathrm{JS}$) across coding and writing tasks. $\uparrow$ indicates higher is better and $\downarrow$ indicates lower is better. Darker blue indicates closer behavioral distribution to real users. Subscripts show the change relative to the better of the two individual simulators}
  \vspace{1em}
  \label{tab:reducing_the_gap}
  \centering
  \resizebox{\textwidth}{!}{%
  \begin{tabular}{l ccc ccc}
    \toprule
    & \multicolumn{3}{c}{\textbf{Coding}} & \multicolumn{3}{c}{\textbf{Writing}} \\
    \cmidrule(lr){2-4} \cmidrule(lr){5-7}
    \textbf{User Simulator}
      & $\mathrm{KL_{\mathrm{fwd}}}$\,$\downarrow$
      & $\mathrm{KL_{\mathrm{bwd}}}$\,$\downarrow$
      & $\mathrm{JS}$\,$\downarrow$
      & $\mathrm{KL_{\mathrm{fwd}}}$\,$\downarrow$
      & $\mathrm{KL_{\mathrm{bwd}}}$\,$\downarrow$
      & $\mathrm{JS}$\,$\downarrow$ \\
    \midrule
    \rowcolor{okabe_teal!20}  \multicolumn{7}{c}{\textit{User Simulator Combinations}} \\
    \midrule

    \texttt{Gemini 3.1 Pro} $+$ \texttt{GPT-5.4}
            & \ccol{56B4E9} .236$_{\downarrow.024}$ & \ccol{56B4E9} .236$_{\downarrow.020}$ & \ccol{56B4E9} .226$_{\downarrow.020}$
            & \ccol{CEE9F9} .462$_{\downarrow.020}$ & \ccol{CEE9F9} .470$_{\downarrow.001}$ & \ccol{C1E3F7} .459$_{\downarrow.007}$ \\
    \texttt{gpt-oss-120b} $+$ \texttt{gemma-4-31B-it}
            & \ccol{EDF7FD} .305$_{\downarrow.006}$ & \ccol{CEE9F9} .294$_{\downarrow.002}$ & \ccol{CEE9F9} .288$_{\downarrow.009}$
            & \ccol{F6FBFE} .514$_{\downarrow.018}$ & \ccol{F6FBFE} .522$_{\downarrow.025}$ & \ccol{F6FBFE} .514$_{\downarrow.023}$ \\
    \texttt{Gemini 3.1 Pro} $+$ \texttt{UserLM-8b}
            & \ccol{8BCCF0} .244$_{\downarrow.016}$ & \ccol{8BCCF0} .269$_{\uparrow.013}$ & \ccol{8BCCF0} .251$_{\uparrow.005}$
            & \ccol{8BCCF0} .405$_{\uparrow.013}$ & \ccol{56B4E9} .415$_{\downarrow.009}$ & \ccol{56B4E9} .401$_{\downarrow.010}$ \\
    \texttt{GPT-5.4} $+$ \texttt{UserLM-8b}
            & \ccol{A2D6F3} .246$_{\downarrow.015}$ & \ccol{A2D6F3} .272$_{\uparrow.007}$ & \ccol{A2D6F3} .257$_{\uparrow.009}$
            & \ccol{56B4E9} .402$_{\uparrow.010}$ & \ccol{8BCCF0} .422$_{\downarrow.002}$ & \ccol{56B4E9} .401$_{\downarrow.010}$ \\
    \texttt{gemma-4-31B-it} $+$ \texttt{UserLM-8b}
            & \ccol{C1E3F7} .269$_{\downarrow.042}$ & \ccol{C1E3F7} .286$_{\downarrow.010}$ & \ccol{C1E3F7} .270$_{\downarrow.027}$
            & \ccol{B3DDF5} .437$_{\uparrow.045}$ & \ccol{C1E3F7} .462$_{\uparrow.038}$ & \ccol{B3DDF5} .441$_{\uparrow.030}$ \\
    \texttt{gpt-oss-120b} $+$ \texttt{UserLM-8b}
            & \ccol{E3F3FB} .303$_{\downarrow.025}$ & \ccol{EDF7FD} .330$_{\downarrow.023}$ & \ccol{EDF7FD} .309$_{\downarrow.034}$
            & \ccol{A2D6F3} .429$_{\uparrow.037}$ & \ccol{A2D6F3} .450$_{\uparrow.026}$ & \ccol{A2D6F3} .438$_{\uparrow.027}$ \\
    \texttt{Gemini 3.1 Pro} $+$ \texttt{GPT-5.4 mini}
            & \ccol{B3DDF5} .266$_{\uparrow.006}$ & \ccol{A2D6F3} .272$_{\uparrow.016}$ & \ccol{B3DDF5} .264$_{\uparrow.018}$
            & \ccol{C1E3F7} .459$_{\downarrow.023}$ & \ccol{B3DDF5} .461$_{\downarrow.010}$ & \ccol{CEE9F9} .463$_{\downarrow.003}$ \\
    \texttt{GPT-5.4} $+$ \texttt{GPT-5.4 mini}
            & \ccol{CEE9F9} .292$_{\uparrow.031}$ & \ccol{E3F3FB} .303$_{\uparrow.038}$ & \ccol{D9EEFA} .291$_{\uparrow.043}$
            & \ccol{E3F3FB} .492$_{\uparrow.010}$ & \ccol{D9EEFA} .499$_{\uparrow.010}$ & \ccol{D9EEFA} .491$_{\uparrow.012}$ \\
    \texttt{gemma-4-31B-it} $+$ \texttt{GPT-5.4 mini}
            & \ccol{D9EEFA} .298$_{\downarrow.013}$ & \ccol{D9EEFA} .298$_{\uparrow.002}$ & \ccol{E3F3FB} .295$_{\downarrow.002}$
            & \ccol{EDF7FD} .504$_{\downarrow.009}$ & \ccol{EDF7FD} .512$_{\downarrow.021}$ & \ccol{EDF7FD} .507$_{\downarrow.013}$ \\
    \texttt{gpt-oss-120b} $+$ \texttt{GPT-5.4 mini}
            & \ccol{F6FBFE} .328$_{\downarrow.030}$ & \ccol{F6FBFE} .339$_{\downarrow.014}$ & \ccol{F6FBFE} .325$_{\downarrow.018}$
            & \ccol{D9EEFA} .491$_{\downarrow.022}$ & \ccol{E3F3FB} .510$_{\downarrow.023}$ & \ccol{E3F3FB} .499$_{\downarrow.021}$ \\
    \texttt{gemma-4-31B-it} $+$ \texttt{Claude Haiku 4.5}
            & \ccol{FFFFFF} .364$_{\uparrow.053}$ & \ccol{FFFFFF} .345$_{\uparrow.049}$ & \ccol{FFFFFF} .346$_{\uparrow.049}$
            & \ccol{FFFFFF} .562$_{\uparrow.025}$ & \ccol{FFFFFF} .568$_{\uparrow.021}$ & \ccol{FFFFFF} .572$_{\uparrow.027}$ \\
    \bottomrule
  \end{tabular}}
\end{table*}

Since our pairwise analysis shows that some user simulators behave distinctly from each other, we test whether combining pairs can reduce the gap to real users by randomly sampling which simulator generates each conversation. Table~\ref{tab:reducing_the_gap} reports the results.

First, we pair the simulators with the smallest distributional gap to real users: \texttt{Gemini 3.1 Pro} with \texttt{GPT-5.4}, and \texttt{gpt-oss-120b} with \texttt{gemma-4-31B-it}. Both pairs achieve smaller $\mathrm{JS}$ divergence than either individual simulator on coding and writing. Next, we pair simulators that are behaviorally distinct from others, \texttt{UserLM-8b} and \texttt{GPT-5.4 mini}, with each of \texttt{Gemini 3.1 Pro}, \texttt{GPT-5.4}, \texttt{gemma-4-31B-it}, and \texttt{gpt-oss-120b}. These improve over the individual simulators on writing and remain competitive on coding. Conversely, combining behaviorally similar simulators, such as \texttt{gemma-4-31B-it} and \texttt{Claude Haiku 4.5}, does not help. 

These results suggest that combining behaviorally complementary simulators can reduce the distributional gap to real users. This combination strategy is conceptually similar to mixture-of-experts \citep{shazeer2017, JMLR:v23:21-0998} and assistant-side routing \citep{peng-etal-2017-composite, wang2025mixtureofagents, ong2025routellm}, and learning a router that selects the appropriate simulator is a natural extension we leave to future work.

\subsection{Interpreting the Behavioral Clusters}
\label{sec:discussion_interpretation}

\begin{figure*}[h]
    \centering
    \includegraphics[width=\textwidth]{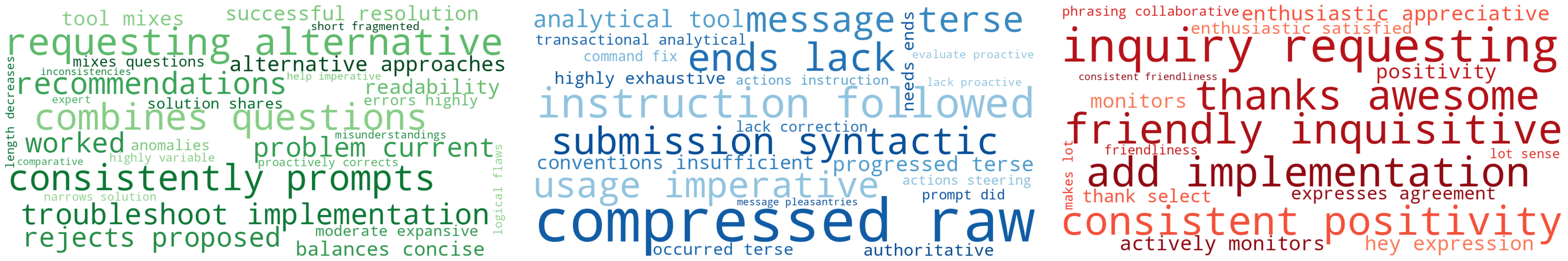}
    \caption{Word clouds of \textit{well-captured} (green), \textit{missed} (blue), and \textit{hallucinated} (red) behaviors for \texttt{Gemini 3.1 Pro} on coding based on a TF-IDF analysis on behavioral clusters.}
    \label{fig:wordcloud_well_captured}
    \vspace{-0.5em}
\end{figure*}

The behavioral distributions reveal not only how much simulators diverge from real users, but which behaviors they capture, miss, or hallucinate. We classify each cluster by the ratio of real to simulated conversations into three categories: \textit{well-captured} if real and simulated are roughly balanced, \textit{missed} if real users dominate, and \textit{hallucinated} if simulated users dominate. For each category, we extract distinctive terms from the behavior descriptions in the top 100 clusters using TF-IDF over unigrams and bigrams. Each term is scored by its TF-IDF in the target category minus the average TF-IDF over the other two. To reduce noise, we filter stopwords and meaningless terms (identified by an LLM), and require terms to appear in at least 50 descriptions.

We present our analysis for \texttt{Gemini 3.1 Pro} on the coding task. Figure~\ref{fig:wordcloud_well_captured} shows the word clouds for each category, and Appendix~\ref{appendix:interpreting_clusters} presents the full table of top 50 terms for each category. We observe the following across the three categories: \textit{Well-captured:} terms reflect iterative problem-solving behaviors such as requesting alternatives, rejecting proposed solutions, and proactively identifying errors (e.g., "requesting alternative", "rejects proposed", "proactively corrects"); \textit{Missed:} terms reflect behaviors the simulator fails to reproduce: terse and compressed communication, transactional exchanges, authoritative directives (e.g., "message terse", "authoritative", "transactional analytical"); \textit{Hallucinated:} terms reflect overly enthusiastic behaviors such as excessive positivity and gratitude, and frequent greetings and farewells (e.g., "friendliness", "thanks awesome", "formal pleasantries").
\section{Related Work}

\paragraph{User Simulation.} LLMs have demonstrated the ability to simulate believable human behavior, and are increasingly used as user simulators for interactive evaluation \citep{davidson2023usersimulationlargelanguage, 10.1145/3596510, zhou2024sotopia, barres2025tau2benchevaluatingconversationalagents, laban2026llms}, synthetic data generation \citep{ding-etal-2023-enhancing, ge2025scalingsyntheticdatacreation, prabhakar2026apigenmt}, and reinforcement learning \citep{zhou2024archer, wu2025collabllm}. Across these applications, initial approaches rely on prompting LLMs \citep{davidson2023usersimulationlargelanguage, park2023generative, park2024generative, luo-etal-2024-duetsim, prabhakar2026apigenmt}. However, these prompt-based simulators struggle to consistently adhere to assigned user goals or profiles, motivating recent methods that train LLM-based user simulators with goal-tracking or persona-consistency objectives \citep{mehri2025goal, abdulhai2026consistently}. In order to build simulators that can faithfully represent real users, other works train LLMs with response-level similarity objectives, such as supervised fine-tuning on real user data \citep{naous2026flipping} or aligning latent reasoning states through reinforcement learning \citep{gandhi2026learning, wu2026humanlm, zhang2026userlmr1modelinghumanreasoning}.

\paragraph{User Simulator Evaluation.} Recent works investigate how LLM-based user simulators compare to real users. One direction compares assistant evaluations conducted with simulators against those conducted with real users, finding that simulator-based evaluations are poor estimates of agent performance and obscure demographic disparities \citep{dou-etal-2025-simulatorarena, seshadri2026lostsimulationllmsimulatedusers}. A second direction compares simulator and human behaviors at the feature level, finding systematic divergence on lexical, syntactic, and stylistic metrics \citep{ivey2024realroboticassessingllms, zhou2026mind, chen2026realworldhumanbehaviorsimulation}.

Across both training and evaluation work, simulator fidelity is typically measured at the response or feature level. While some works investigate simulating diverse user populations \citep{liu2023one, pmlr-v202-aher23a, ge2025scalingsyntheticdatacreation}, ours is the first to measure how well user simulators capture the distribution of real user behaviors. We do this by extracting representations of user behavior that go beyond lexical or stylistic features, and clustering them to compare the distributions of real and simulated user behaviors.

\paragraph{Distributional Metrics.} Distributional metrics for generative models compare generated and real distributions through divergence-based measures and precision-recall metrics, originally developed for image generation \citep{NEURIPS2018_f7696a9b, NEURIPS2019_0234c510, pmlr-v108-djolonga20a}. Subsequent work has extended these methods to text generation, developing divergence-based metrics over quantized distributions of generated and real text \citep{pillutla2021mauve, JMLR:v24:23-0023, pimentel2023on}. Our work further extends these methods to user simulation, comparing the distributions of user behaviors among real and simulated users.
\section{Conclusion}
\label{sec:conclusion}

We introduce a method for measuring how well user simulators represent the broad, heterogeneous distribution of real user behaviors. Through a systematic evaluation of 24 LLM-based user simulators across coding and writing tasks, we reveal a large distributional gap from real users and surface interpretable behavioral patterns that simulators capture, miss, and hallucinate. We take initial steps towards closing this gap by combining pairs of behaviorally complementary simulators. We hope this work motivates future research toward user simulators that faithfully represent the diverse behaviors of real-world users.

\clearpage
\newpage

\bibliography{custom}
\bibliographystyle{apalike}

\clearpage
\newpage

\appendix

\section{Limitations and Future Directions}
\label{appendix:limitations}

We make an effort to approximate the distribution of real user behaviors by grounding our evaluation in 10,000 real-world conversations between users and assistants across coding and writing tasks. However, there are some limitations to how well this references reflects user behaviors in the real world. The conversations are drawn from a single source, WildChat \citep{zhao2024wildchat}, which was collected through a public chatbot interface hosted on Hugging Face Spaces. This likely skews the user population toward the IT/developer community and may not adequately represent other populations who interact with LLMs in different settings. The evaluation also focuses on English conversations, which potentially underrepresents the linguistic, cultural, and regional diversity of user interactions with LLMs across the world. Moreover, our analysis does not extend to tasks other than coding or writing, where user behaviors may differ.

Our method measures the distributional gap between real and simulated user behaviors by comparing the behaviors demonstrated in real and simulated conversations. One limitation of our approach is that real conversations had ChatGPT/GPT-4 as the assistants, while our simulated conversations were generated using Qwen3.5-122B-A10B as the assistant. The way that users behave is partly shaped by the assistant, so the measured gap may be affected by differences assistant behavior. Additionally, simulated conversations were generated based on user goals that were extracted from real conversations using GPT-4o. The user goal extraction process can be lossy, stripping context the original user had. If user simulators were provided with this context, such as personas or prior experience, it may help address the gap and enable them to more faithfully represent user behaviors.

Through a human study and a series of validation experiments, we confirm that our method captures meaningful behavioral distributions and is robust to the choice of embedding model and clustering algorithm. We would like acknowledge, however, that our approach relies on a capable LLM (Qwen3.5-122B-A10B-FP8) to analyze conversations and generate accurate descriptions of user behavior.

In our work, we describe that user simulators can fail to capture the distribution of real users when they either demonstrate behaviors that real users rarely exhibit, or when they don't demonstrate behaviors that real users do exhibit. However, in some cases this is actually desirable. For example, when user simulators are used to stress-test assistants against rare or adversarial user inputs that the real-world data does not capture well. We do not distinguish between desirable and undesirable divergence, and leave this for future work. Lastly, our work focuses more on intrinsic analysis, and does not measure how user simulator distributional gaps affect downstream training or evaluation for assistants.

User simulators serve an important role in the development process of AI systems, often being used during training or evaluation. When simulators underrepresent the behaviors of certain user populations, these assistant may fail to serve those users well, and the burden of this failure falls on the populations whose behaviors are underrepresented. Closing this distributional gap is therefore important for building AI systems that work for users all over the world, including different cultures, languages, and demographics. Beyond this, high-fidelity user simulators raise dual-use concerns and can be misused to impersonate real users or create fake users for unwanted purposes. Addressing these limitations is important for advancing user simulators that faithfully represent real users and support the responsible development of AI systems.

\clearpage
\newpage
\section{User Behavior Representation Prompts}
\label{appendix:user_behavior_representations}

\begin{figure}[H]
\begin{tcolorbox}[
    colback=gray!5,
    colframe=gray!75!black,
    title=User Behavior Representation Prompt: Requests Facet,
    fonttitle=\bfseries,
    left=2mm,
    right=2mm,
    top=1mm,
    bottom=1mm,
    arc=2mm,
    boxrule=0.5pt,
    breakable
]
\begin{lstlisting}[
    basicstyle=\scriptsize\ttfamily,
    breaklines=true,
]
You are an expert analyzing user behaviour in human-AI conversations. The user has a goal, and the assistant helps them achieve it. Your task is to describe the user's behavior according to the criteria below.

# User Goal
{user_goal}
# User Utterances
{conversation_history}

# Analysis Criteria
1. Specification and Articulation
   - How specified are the requests? Is the first request underspecified, and clarified in subsequent turns? Or are the requests exhaustive? 
   - What type of information is left underspecified/specified? (e.g. constraints, edge cases, context, output format, etc.)
   - Does the user explicitly command specific tasks, or do they rely on indirect cues (e.g., presenting an error without explicitly asking for a fix)?

2. User Goal Decomposition
   - How is the user goal decomposed across utterances?
      - Single-shot: The entire goal is expressed in one utterance with no further decomposition.
      - Top-down: High-level goal stated upfront, then progressively refined or elaborated.
      - Bottom-up: Individual preconditions or sub-tasks addressed first, building toward the overall goal.
      - Chained: Each request builds purely on the immediate previous turn rather than referring back to a central goal.

3. Relevance to Goal
   - Are the requests directly related to the user goal? Or does the user introduce secondary/perpendicular/emergent needs? 
   - What functions do the requests serve beyond achieving the user goal? (setting context, probing AI capabilities, verifying intermediate outputs, logistics, troubleshooting, side-effects, exploring related sub-tasks, etc.)

# Instructions
- Generate terse, concise bullet points, not full sentences.
- Abstract away from the specific topic/domain
- IMPORTANT:Do NOT use task-specific terms (e.g., "coding," "booking," "Python", "CSV"). Use generic substitutes (e.g., "executing a task," "providing constraints," "the target artifact").

Output a valid JSON object using the exact format below. Do not include any text outside the JSON.

{{
   "specification_and_articulation": {{
      "specification_level": "how specified are requests?",
      "underspecified_aspects": "what types of information are left underspecified/specified? (provide high-level descriptions without task-specific details)",
      "articulation_mode": "how does the user articulate their requests (explicit directives, indirect cues, mixed)?"
   }},
   "goal_decomposition_strategy": "describe how the user goal decomposed across utterances?",
   "relevance_to_goal": {{
      "goal_adherence": "Are the user's requests directly related to the user goal? If not, what other functions do they serve? (provide high-level descriptions without task-specific details)"
   }}
}}
\end{lstlisting}
\end{tcolorbox}
\caption{Prompt for extracting user behavior representations along the \textit{Requests} facet.}
\label{fig:prompt_req_dim}
\end{figure}

\newpage
\begin{figure}[H]
\begin{tcolorbox}[
    colback=gray!5,
    colframe=gray!75!black,
    title=User Behavior Representation Prompt: Responses Facet,
    fonttitle=\bfseries,
    left=2mm,
    right=2mm,
    top=1mm,
    bottom=1mm,
    arc=2mm,
    boxrule=0.5pt,
    breakable
]
\begin{lstlisting}[
    basicstyle=\scriptsize\ttfamily,
    breaklines=true,
]
You are an expert analyzing user behaviour in human-AI conversations. The user has a goal, and the assistant helps them achieve it. Your task is to describe the user's behavior according to the criteria below.

# User Goal
{user_goal}

# Conversation
{conversation_history}

# Analysis Criteria
1. Engagement and Evaluation
   - Does the user engage with the agent responses, or ignore/skip over them?
   - How does the user evaluate the agent's output? (e.g., explicit validation, implicit acceptance by continuing, partial acceptance with corrections, rejection)
   - Does the user provide specific or actionable feedback, or only surface-level acknowledgment?

2. Response Composition
   - What types of actions are present in the user's responses? Are they reactive (e.g., validation, acknowledgment, answering agent questions, corrections, feedback), proactive (e.g., follow-up requests, new constraints/preferences, suggestions, questions), or self-directed (e.g., thinking out loud, expressing uncertainty, narrating their process)?

3. Steering Mechanism
   - Does the user steer through direct follow-up requests, or through indirect means? (e.g., asking questions that implicitly request action, expressing dissatisfaction without stating what to change, providing hints or examples rather than directives)
   - Does the user introduce new preferences, constraints, or feedback as part of their responses, effectively reshaping the task incrementally?

# Instructions
- Generate terse, concise bullet points, not full sentences.
- Abstract away from the specific topic/domain.
- IMPORTANT: Do NOT use task-specific terms (e.g., "coding," "booking," "Python", "CSV"). Use generic substitutes (e.g., "executing a task," "providing constraints," "the target artifact").

Output a valid JSON object using the exact format below. Do not include any text outside the JSON.

{{
   "engagement_and_evaluation": {{
      "engagement_level": "does the user engage with the agent's responses or skip over them?",
      "evaluation_mode": "how does the user evaluate the agent's output? (explicit validation, implicit acceptance, partial acceptance, rejection, etc.)",
      "feedback_specificity": "does the user provide specific/actionable feedback or only surface-level acknowledgment?"
   }},
   "response_composition": {{
      "action_types": "What types of actions are present in the user's responses? Are they reactive (e.g., validation, acknowledgment, answering agent questions, corrections, feedback), proactive (e.g., follow-up requests, new constraints/preferences, suggestions, questions), or self-directed (e.g., thinking out loud, expressing uncertainty, narrating their process)?"
   }},
   "steering_mechanism": {{
      "directness": "does the user steer through explicit follow-up requests or indirect means (questions, hints, expressed dissatisfaction)?",
      "incremental_reshaping": "does the user introduce new preferences/constraints/feedback that reshape the task within their responses?"
   }}
}}
\end{lstlisting}
\end{tcolorbox}
\caption{Prompt for extracting user behavior representations along the \textit{Responses} facet.}
\label{fig:prompt_resp_dim}
\end{figure}

\newpage
\begin{figure}[H]
\begin{tcolorbox}[
    colback=gray!5,
    colframe=gray!75!black,
    title=User Behavior Representation Prompt: Context Facet,
    fonttitle=\bfseries,
    left=2mm,
    right=2mm,
    top=1mm,
    bottom=1mm,
    arc=2mm,
    boxrule=0.5pt,
    breakable
]
\begin{lstlisting}[
    basicstyle=\scriptsize\ttfamily,
    breaklines=true,
]
You are an expert analyzing user behaviour in human-AI conversations. The user has a goal, and the assistant helps them achieve it. Your task is to describe the user's behavior according to the criteria below.

# User Goal
{user_goal}

# Conversation
{conversation_history}

# Analysis Criteria
1. Context Richness
   - How much context does the user provide overall? Do they share background about themselves, their situation, what they've already tried, what they're struggling with, or what they're thinking?
   - Does the user provide all relevant context needed for the agent to help effectively, or is essential context left out?
   - Does the user provide context at all, or do they simply issue directives without situating the task?

2. Context Type
   - What types of context does the user provide? (e.g., personal background, goals/motivations, prior attempts, existing solutions, constraints, preferences, domain knowledge, emotional state, thought process)

3. Context Delivery
   - Does the user front-load all relevant context in their first message, or reveal it incrementally across turns? What types of context are introduced later? (e.g., constraints they forgot, preferences they didn't think to mention, background that becomes relevant as the task evolves)
   - Is incremental context revealed in response to agent questions, or volunteered unprompted?

# Instructions
- Generate terse, concise bullet points, not full sentences.
- Abstract away from the specific topic/domain.
- IMPORTANT: Do NOT use task-specific terms (e.g., "coding," "booking," "Python", "CSV"). Use generic substitutes (e.g., "executing a task," "providing constraints," "the target artifact").

Output a valid JSON object using the exact format below. Do not include any text outside the JSON.

{{
   "context_richness": {{
      "depth": "how much context does the user provide? do they share background, prior attempts, thought process, etc.?",
      "completeness": "does the user provide all relevant context or leave essential information out?",
      "contextualization_vs_directives": "does the user situate the task with context, or simply issue directives?"
   }},
   "context_type": "what types of context does the user provide? (personal background, goals/motivations, prior attempts, existing solutions, constraints, preferences, domain knowledge, emotional state, thought process, etc.)",
   "context_delivery": {{
      "distribution": "does the user front-load context or reveal it incrementally? What types of context are introduced in later turns?",
      "trigger": "is incremental context revealed in response to agent questions or volunteered unprompted?"
   }}
}}
\end{lstlisting}
\end{tcolorbox}
\caption{Prompt for extracting user behavior representations along the \textit{Context} facet.}
\label{fig:prompt_context_dim}
\end{figure}

\newpage
\begin{figure}[H]
\begin{tcolorbox}[
    colback=gray!5,
    colframe=gray!75!black,
    title=User Behavior Representation Prompt: Communication Style Facet,
    fonttitle=\bfseries,
    left=2mm,
    right=2mm,
    top=1mm,
    bottom=1mm,
    arc=2mm,
    boxrule=0.5pt,
    breakable
]
\begin{lstlisting}[
    basicstyle=\scriptsize\ttfamily,
    breaklines=true,
]
You are an expert analyzing user behaviour in human-AI conversations. The user has a goal, and the assistant helps them achieve it. Your task is to describe the user's behavior according to the criteria below.

# User Goal
{user_goal}

# Conversation
{conversation_history}

# Analysis Criteria
1. Register and Tone
   - What is the user's register? (e.g., formal, casual, conversational, terse, professional)
   - What is the user's emotional tone? (e.g., neutral, frustrated, enthusiastic, impatient, apologetic, deferential)
   - Does the register/tone shift across the conversation? (e.g., starts polite but becomes curt after errors)

2. Verbosity and Structure
   - How verbose are the user's messages? Are they minimal and compressed, or expansive and elaborated?
   - Does the user use formatting conventions? (e.g., bullet points, numbered lists, code blocks, markdown, all caps, punctuation patterns)
   - How does message length and structure change across turns?

3. Social Conventions
   - Does the user use greetings, pleasantries, expressions of gratitude, or sign-offs?
   - Does the user treat the agent as a tool (purely transactional) or as a social interlocutor (rapport-building, politeness, humor)?

4. Request Framing
   - How does the user syntactically frame their requests? (e.g., imperative commands, questions, hedged suggestions, statements of problems without explicit asks)

# Instructions
- Generate terse, concise bullet points, not full sentences.
- Abstract away from the specific topic/domain.
- IMPORTANT: Do NOT use task-specific terms (e.g., "coding," "booking," "Python", "CSV"). Use generic substitutes (e.g., "executing a task," "providing constraints," "the target artifact").

Output a valid JSON object using the exact format below. Do not include any text outside the JSON.

{{
   "register_and_tone": {{
      "register": "what is the user's register? (formal, casual, conversational, terse, professional, etc.)",
      "emotional_tone": "what is the user's emotional tone? (neutral, frustrated, enthusiastic, impatient, apologetic, deferential, etc.)",
      "tone_shifts": "does the tone shift across the conversation, and in response to what?"
   }},
   "verbosity_and_structure": {{
      "verbosity": "how verbose are the user's messages? minimal and compressed, or expansive and elaborated?",
      "formatting": "does the user use formatting conventions? (bullet points, numbered lists, code blocks, markdown, etc.)",
      "evolution": "how does message length and structure change across turns?"
   }},
   "social_conventions": {{
      "politeness_markers": "does the user use greetings, pleasantries, gratitude, or sign-offs?",
      "agent_relationship": "does the user treat the agent as a tool (transactional) or as a social interlocutor (rapport-building, politeness, humor)?"
   }},
   "request_framing": "how does the user syntactically frame their requests? (imperative commands, questions, hedged suggestions, problem statements without explicit asks, etc.)"
}}
\end{lstlisting}
\end{tcolorbox}
\caption{Prompt for extracting user behavior representations along the \textit{Communication Style} facet.}
\label{fig:prompt_comm_style_dim}
\end{figure}

\newpage
\begin{figure}[H]
\begin{tcolorbox}[
    colback=gray!5,
    colframe=gray!75!black,
    title=User Behavior Representation Prompt: DAMSL Dialog Acts Facet,
    fonttitle=\bfseries,
    left=2mm,
    right=2mm,
    top=1mm,
    bottom=1mm,
    arc=2mm,
    boxrule=0.5pt,
    breakable
]
\begin{lstlisting}[
    basicstyle=\scriptsize\ttfamily,
    breaklines=true,
]
You are an expert analyzing user behaviour in human-AI conversations. Your task is to analyze a specific user utterance in a multi-turn conversation and provide a description of the user's behaviour.

# Conversation History
{conversation_history}

# Target Utterance
Read the conversation history for context, but your analysis MUST strictly apply ONLY to this utterance:
{target_user_utterance}


# Dialogue Act Markup in Several Layers (DAMSL) Annotation Framework
DAMSL defines utterances based on the intentions of the speaker. Each utterance is analyzed along several dimensions.
1. INFORMATION LEVEL: Characterizes the semantic content of the utterance:
    - **Task**: Directly advances (or attempts to advance) the goals of the domain task.
    - **Task-management**: Addresses the problem-solving process/procedure rather than performing the task itself.
    - **Communication-management**: Concerned exclusively with maintaining the communication channel.
2. FORWARD LOOKING FUNCTION: Characterizes what effect the utterance has on the subsequent dialog:
    - **Statement**: Makes a claim (Assert, Re-assert, Other-statement).
    - **Info-request**: Asks for information.
    - **Influencing-addressee-future-action**: Directly influences the addressee's future non-communicative actions, directs or suggests the addressee to perform an action.
    - **Committing-speaker-future-action**: Potentially commits the speaker to some future action.
    - **Conventional**: Conventional conversational actions such as greeting, farewells, thanking, or responding to thanks.
3. BACKWARD LOOKING FUNCTION: Characterizes how the current utterance relates to the previous discourse:
    - **Agreement**: Accept, reject, partial accept/reject, holds off a response.
    - **Answer**: Provides requested information.
    - **Understanding**: Signals comprehension (acknowledgement, repetition/reformulation, collaborative completion) or lack thereof.


# Output Instructions
Analyze the target utterance along DAMSL's dimensions and write a concise, 1 sentence description of the user's behaviour and its functional role in the target utterance:
- Generate terse, concise bullet points, not full sentences.
- Abstract away from the specific topic/domain
- IMPORTANT:Do NOT use task-specific terms (e.g., "coding," "booking," "Python", "CSV"). Use generic substitutes (e.g., "executing a task," "providing constraints," "the target artifact").

Output only the concise description as plain text, without any other text or formatting.
\end{lstlisting}
\end{tcolorbox}
\caption{Prompt for extracting user behavior representations along the \textit{DAMSL Dialog Acts} facet.}
\label{fig:prompt_damsl_dim}
\end{figure}

\newpage
\begin{figure}[H]
\begin{tcolorbox}[
    colback=gray!5,
    colframe=gray!75!black,
    title=User Behavior Representation Prompt: SGD Dialog Acts Facet,
    fonttitle=\bfseries,
    left=2mm,
    right=2mm,
    top=1mm,
    bottom=1mm,
    arc=2mm,
    boxrule=0.5pt,
    breakable
]
\begin{lstlisting}[
    basicstyle=\scriptsize\ttfamily,
    breaklines=true,
]
You are an expert analyzing user behaviour in human-AI conversations. Your task is to analyze a specific user utterance in a multi-turn conversation and classify the user's dialogue act(s).

# Conversation History
{conversation_history}

# Target Utterance
Read the conversation history for context, but your analysis MUST strictly apply ONLY to this utterance:
{target_user_utterance}


# Dialogue Act Taxonomy (adapted from Schema-Guided Dialogue)
Classify the target utterance using one or more of the following user dialogue acts:

1. INFORM: User provides information, states preferences, or supplies constraints to the agent.
2. REQUEST: User asks the agent for specific information or details.
3. AFFIRM: User agrees with, confirms, or accepts a proposition, value, or suggestion made by the agent.
4. NEGATE: User disagrees with, denies, or rejects a proposition, value, or suggestion made by the agent.
5. SELECT: User chooses between multiple options or alternatives presented by the agent.
6. INFORM_INTENT: User states, introduces, or shifts to a new goal or task objective.
7. AFFIRM_INTENT: User confirms or agrees with a goal or task objective suggested by the agent.
8. NEGATE_INTENT: User rejects or denies a goal or task objective suggested by the agent.
9. REQUEST_ALTS: User asks for different or additional options beyond what has been presented.
10. THANK: User expresses gratitude or appreciation.
11. GOODBYE: User signals the end of the conversation or a conversational closing.
12. GREET: User initiates the conversation with a greeting or opening.


# Output Instructions
- Assign ALL applicable dialogue acts to the target utterance (utterances may have multiple acts).

Output the dialogue acts, separated by commas, without any other text or formatting.
\end{lstlisting}
\end{tcolorbox}
\caption{Prompt for extracting user behavior representations along the \textit{SGD Dialog Acts} facet.}
\label{fig:prompt_sgd_dim}
\end{figure}

\newpage
\section{User Goal Classification}
\label{appendix:user_goal_classification}

\begin{figure}[H]
\begin{tcolorbox}[
    colback=gray!5,
    colframe=gray!75!black,
    title=User Goal Classification Prompt,
    fonttitle=\bfseries,
    left=2mm,
    right=2mm,
    top=1mm,
    bottom=1mm,
    arc=2mm,
    boxrule=0.5pt,
    breakable
]
\begin{lstlisting}[
    basicstyle=\scriptsize\ttfamily,
    breaklines=true,
]
You are an expert classifier. Given a user intent, your task is to classify it into one category and one subcategory.

# User Intent
{user_intent}

## Categories and Subcategories

1. **Artifact Creation** - The user wants to produce a final artifact (code, writing, prompt, etc.).
    - Subcategories: Writing, Coding, Prompt Generation, Other
2. **Information Seeking** - The user wants to receive information about a topic.
    - Subcategories: Writing, Coding, Math, Science, Other
3. **Practical Guidance** - includes activities like tutoring and teaching, how-to advice about a variety of topics, and creative ideation (highly customized to the user and can be adapted based on conversation and follow-up)
    - Subcategories: Teaching, How-To Advice, Self-Care, Creative Ideation, Other
4. **Other** - The intent does not clearly fit any of the above categories.
    - Subcategories: Other

## Output Format

For each response, output a valid JSON object using the exact format below. Use double quotes ("), escape any double quotes within strings using backslashes (\"), escape newlines as \\n, and do not include any text before or after the JSON object.

{{
    "category": str, # One of: "Artifact Creation", "Information Seeking", "Practical Guidance", "Other"
    "subcategory": str # One of: "Writing", "Coding", "Prompt Generation", "Math", "Science", "Teaching", "How-To Advice", "Self-Care", "Creative Ideation", "Other"
}}
\end{lstlisting}
\end{tcolorbox}
\caption{Prompt for classifying user goals.}
\label{fig:prompt_goal_classification}
\end{figure}

\newpage
\section{Conversation Generation Prompts}
\label{appendix:conversation_generation_prompts}

\begin{figure}[H]
\begin{tcolorbox}[
    colback=gray!5,
    colframe=gray!75!black,
    title=Assistant System Prompt,
    fonttitle=\bfseries,
    left=2mm,
    right=2mm,
    top=1mm,
    bottom=1mm,
    arc=2mm,
    boxrule=0.5pt,
    breakable
]
\begin{lstlisting}[
    basicstyle=\scriptsize\ttfamily,
    breaklines=true,
]
You are a helpful assistant. Respond to the user's requests accurately and helpfully.
\end{lstlisting}
\end{tcolorbox}
\caption{Assistant system prompt for conversation generation.}
\label{fig:prompt_assistant}
\end{figure}

\begin{figure}[H]
\begin{tcolorbox}[
    colback=gray!5,
    colframe=gray!75!black,
    title=User Simulator System Prompt,
    fonttitle=\bfseries,
    left=2mm,
    right=2mm,
    top=1mm,
    bottom=1mm,
    arc=2mm,
    boxrule=0.5pt,
    breakable
]
\begin{lstlisting}[
    basicstyle=\scriptsize\ttfamily,
    breaklines=true,
]
You are a user simulator interacting with an AI assistant to achieve a goal.

# User Goal
{intent}

# Instructions
- Users can make typos, they don't always use perfect punctuation, and they tend to be lazy because typing requires effort.
- You should split information across turns and not give everything at the start.
- However, do not overdo any of these things - you must realistically act like a human.
- If your goal has been fulfilled, respond ONLY with \"{termination_signal}\".
\end{lstlisting}
\end{tcolorbox}
\caption{User simulator system prompt for conversation generation.}
\label{fig:prompt_user_simulator}
\end{figure}

\section{Base vs. Post-Trained Simulators}
\label{appendix:base_vs_posttrained}

To take an initial step towards understanding how post-training affects simulating user behaviors, we conduct a preliminary experiment comparing \texttt{Qwen3.5-4B-Base} with its post-trained counterpart \texttt{Qwen3.5-4B} on the writing task. We follow the same setup described in Section~\ref{sec:experimental_setup}. Table~\ref{tab:base_vs_posttrained} reports the results. The base model achieves lower divergence than the post-trained model across all three metrics. As this experiment covers a single model pair on a single task, we view these results as preliminary, and leave further investigation of how post-training affects the ability of LLMs to simulate diverse user behaviors to future work.

\begin{table}[h!]
    \caption{
    The distributional gap between real and simulated user behaviors for a base model and its post-trained counterpart on the writing task. We compare the distribution of user behaviors demonstrated in $\mathcal{D}_{\text{real}}$ with those in $\mathcal{D}_{\text{sim}}$. $\uparrow$ indicates higher is better and $\downarrow$ indicates lower is better.
    }
  \label{tab:base_vs_posttrained}
  \centering
  \begin{tabular}{l ccc}
    \toprule
    & \multicolumn{3}{c}{\textbf{Writing}} \\
    \cmidrule(lr){2-4}
    \textbf{User Simulator}
      & $\mathrm{KL_{\mathrm{fwd}}}$\,$\downarrow$
      & $\mathrm{KL_{\mathrm{bwd}}}$\,$\downarrow$
      & $\mathrm{JS}$\,$\downarrow$ \\
    \specialrule{0.9pt}{2pt}{2pt}

    \texttt{Qwen3.5-4B-Base}
            & 0.554 & 0.555 & 0.557 \\
    \texttt{Qwen3.5-4B}
            & 0.597 & 0.609 & 0.609 \\

    \bottomrule
  \end{tabular}
\end{table}

\newpage
\section{Results for Each User Behavior Facet}
\label{appendix:results_per_facet}

In addition to the metrics defined in Section \ref{sec:method}, all results in the Appendix include two complementary metrics:
\begin{itemize}[leftmargin=12pt,topsep=0pt]
    \item \textbf{Nearest Neighbor Cosine Similarity:} The average cosine similarity between each real user and its nearest neighbor in the simulated set. Higher values indicate that simulated behaviors are closer to real ones in the embedding space. This metric is computed directly on the embeddings.
    \item \textbf{MAUVE:} Compares $\hat{P}$ and $\hat{Q}$ using divergence frontiers \citep{pillutla2021mauve}. The mixture distribution $R_\lambda = \lambda\hat{P} + (1-\lambda)\hat{Q}$ yields the divergence curve $\mathcal{C}(\hat{P},\hat{Q}) = \{(\exp(-c\,\mathrm{KL}(\hat{Q}\,\|\,R_\lambda)),\, \exp(-c\,\mathrm{KL}(\hat{P}\,\|\,R_\lambda))),\, \lambda \in (0,1)\}$. The MAUVE score is the area under this curve, which captures both failure modes. Higher values indicate better alignment between real and simulated users. We clarify that we apply the divergence-frontier formulation of \citet{pillutla2021mauve}, not the off-the-shelf text-generation metric.
\end{itemize}

\label{appendix:per_dimension_results}
\definecolor{okabe_orange}{HTML}{E69F00}
\definecolor{okabe_blue}{HTML}{56B4E9}
\definecolor{okabe_pink}{HTML}{CC79A7}

\begin{table*}[h!]
    \caption{
    Requests facet results. The distributional gap between real and simulated user behaviors across coding and writing tasks. We compare the distribution of user behaviors demonstrated in $\mathcal{D}_{\text{real}}$ with those in $\mathcal{D}_{\text{sim}}$. $\uparrow$ indicates higher is better and $\downarrow$ indicates lower is better.
    }
  \label{tab:main_results_req}
  \centering
  \resizebox{\textwidth}{!}{%
  \begin{tabular}{l ccccc ccccc}
    \toprule
    & \multicolumn{5}{c}{\textbf{Coding}} & \multicolumn{5}{c}{\textbf{Writing}} \\
    \cmidrule(lr){2-6} \cmidrule(lr){7-11}
    \textbf{User Simulator}
      & $\mathrm{NN}$\,$\uparrow$
      & $\mathrm{KL_{\mathrm{fwd}}}$\,$\downarrow$
      & $\mathrm{KL_{\mathrm{bwd}}}$\,$\downarrow$
      & $\mathrm{MAU}$\,$\uparrow$
      & $\mathrm{JS}$\,$\downarrow$
      & $\mathrm{NN}$\,$\uparrow$
      & $\mathrm{KL_{\mathrm{fwd}}}$\,$\downarrow$
      & $\mathrm{KL_{\mathrm{bwd}}}$\,$\downarrow$
      & $\mathrm{MAU}$\,$\uparrow$ 
      & $\mathrm{JS}$\,$\downarrow$ \\
    \specialrule{0.9pt}{2pt}{2pt}
    \rowcolor{okabe_blue} \textbf{\textit{Real Users}}
            & .826 & .034 & .035 & .924 & .039
            & .888 & .024 & .025 & .955 & .029 \\
    \specialrule{0.9pt}{2pt}{2pt}
    \rowcolor{okabe_pink!20}  \multicolumn{11}{c}{\textit{Closed-Source}} \\
    \midrule
    \texttt{GPT-5.4}
            & 0.815 & 0.164 & 0.175 & 0.437 & 0.163
            & 0.862 & 0.337 & 0.355 & 0.111 & 0.330 \\
    \texttt{GPT-5.4 mini}
            & 0.810 & 0.219 & 0.245 & 0.260 & 0.229
            & 0.863 & 0.295 & 0.308 & 0.151 & 0.294 \\
    \texttt{GPT-5.4 nano}
            & 0.804 & 0.269 & 0.291 & 0.198 & 0.262
            & 0.855 & 0.396 & 0.420 & 0.065 & 0.392 \\
    \texttt{Claude Haiku 4.5}
            & 0.798 & 0.293 & 0.289 & 0.171 & 0.279
            & 0.831 & 0.513 & 0.521 & 0.022 & 0.512 \\
    \texttt{Gemini 3.1 Pro}
            & 0.820 & 0.131 & 0.134 & 0.570 & 0.126
            & 0.858 & 0.376 & 0.372 & 0.090 & 0.354 \\
    \texttt{Gemini 3 Flash}
            & 0.808 & 0.228 & 0.211 & 0.308 & 0.208
            & 0.844 & 0.466 & 0.462 & 0.043 & 0.436 \\
    \texttt{Gemini 3.1 Flash-Lite}
            & 0.801 & 0.254 & 0.243 & 0.237 & 0.241
            & 0.845 & 0.456 & 0.454 & 0.040 & 0.445 \\
    \midrule
    \rowcolor{okabe_blue!20} \multicolumn{11}{c}{\textit{Open-Source}} \\
    \midrule
    \texttt{Qwen3.5-122B-A10B}
            & 0.806 & 0.234 & 0.221 & 0.281 & 0.219
            & 0.844 & 0.456 & 0.452 & 0.039 & 0.448 \\
    \texttt{Qwen3.5-35B-A3B}
            & 0.801 & 0.275 & 0.263 & 0.202 & 0.260
            & 0.843 & 0.475 & 0.483 & 0.032 & 0.469 \\
    \texttt{Qwen3.5-27B}
            & 0.798 & 0.280 & 0.262 & 0.201 & 0.260
            & 0.838 & 0.461 & 0.471 & 0.035 & 0.459 \\
    \texttt{Qwen3.5-9B}
            & 0.802 & 0.293 & 0.283 & 0.190 & 0.267
            & 0.843 & 0.473 & 0.478 & 0.035 & 0.461 \\
    \texttt{Qwen3.5-4B}
            & 0.804 & 0.284 & 0.277 & 0.196 & 0.264
            & 0.842 & 0.481 & 0.498 & 0.030 & 0.478 \\
    \texttt{Qwen3.5-2B}
            & 0.786 & 0.416 & 0.421 & 0.056 & 0.408
            & 0.823 & 0.558 & 0.562 & 0.014 & 0.557 \\
    \texttt{Qwen3.5 0.8B}
            & 0.792 & 0.402 & 0.422 & 0.057 & 0.405
            & 0.831 & 0.526 & 0.536 & 0.019 & 0.527 \\
    \texttt{Llama-3.3-70B-Instruct}
            & 0.802 & 0.294 & 0.300 & 0.175 & 0.277
            & 0.831 & 0.500 & 0.512 & 0.024 & 0.502 \\
    \texttt{Llama-3.1-8B-Instruct}
            & 0.800 & 0.299 & 0.284 & 0.186 & 0.270
            & 0.829 & 0.505 & 0.526 & 0.025 & 0.495 \\
    \texttt{gpt-oss-120b}
            & 0.810 & 0.203 & 0.194 & 0.364 & 0.187
            & 0.858 & 0.378 & 0.390 & 0.077 & 0.372 \\
    \texttt{gpt-oss-20b}
            & 0.810 & 0.237 & 0.248 & 0.276 & 0.222
            & 0.851 & 0.461 & 0.484 & 0.037 & 0.455 \\
    \texttt{gemma-4-31B-it}
            & 0.814 & 0.222 & 0.211 & 0.318 & 0.204
            & 0.849 & 0.467 & 0.460 & 0.040 & 0.445 \\
    \texttt{gemma-4-26B-A4B-it}
            & 0.808 & 0.229 & 0.209 & 0.304 & 0.209
            & 0.843 & 0.469 & 0.474 & 0.035 & 0.461 \\
    \texttt{gemma-4-E4B-it}
            & 0.790 & 0.322 & 0.308 & 0.145 & 0.299
            & 0.831 & 0.534 & 0.547 & 0.021 & 0.516 \\
    \texttt{gemma-4-E2B-it}
            & 0.800 & 0.276 & 0.280 & 0.189 & 0.268
            & 0.841 & 0.495 & 0.508 & 0.029 & 0.481 \\
    \midrule
    \rowcolor{okabe_orange!20}\multicolumn{11}{c}{\textit{Trained Simulators}} \\
    \midrule
    \texttt{UserLM-8b}
            & 0.814 & 0.212 & 0.255 & 0.257 & 0.229
            & 0.864 & 0.298 & 0.330 & 0.134 & 0.308 \\
    \texttt{humanlm-opinion}
            & 0.813 & 0.189 & 0.190 & 0.386 & 0.180
            & 0.856 & 0.375 & 0.398 & 0.079 & 0.370 \\
    \bottomrule
  \end{tabular}}
\end{table*}
\definecolor{okabe_orange}{HTML}{E69F00}
\definecolor{okabe_blue}{HTML}{56B4E9}
\definecolor{okabe_pink}{HTML}{CC79A7}

\begin{table*}[h!]
  \caption{
    Responses facet results. The distributional gap between real and simulated user behaviors across coding and writing tasks. We compare the distribution of user behaviors demonstrated in $\mathcal{D}_{\text{real}}$ with those in $\mathcal{D}_{\text{sim}}$. $\uparrow$ indicates higher is better and $\downarrow$ indicates lower is better.
    }
  \label{tab:main_results_response}
  \centering
  \resizebox{\textwidth}{!}{%
  \begin{tabular}{l ccccc ccccc}
    \toprule
    & \multicolumn{5}{c}{\textbf{Coding}} & \multicolumn{5}{c}{\textbf{Writing}} \\
    \cmidrule(lr){2-6} \cmidrule(lr){7-11}
    \textbf{User Simulator}
      & $\mathrm{NN}$\,$\uparrow$
      & $\mathrm{KL_{\mathrm{fwd}}}$\,$\downarrow$
      & $\mathrm{KL_{\mathrm{bwd}}}$\,$\downarrow$
      & $\mathrm{MAU}$\,$\uparrow$
      & $\mathrm{JS}$\,$\downarrow$
      & $\mathrm{NN}$\,$\uparrow$
      & $\mathrm{KL_{\mathrm{fwd}}}$\,$\downarrow$
      & $\mathrm{KL_{\mathrm{bwd}}}$\,$\downarrow$
      & $\mathrm{MAU}$\,$\uparrow$
      & $\mathrm{JS}$\,$\downarrow$ \\
    \specialrule{0.9pt}{2pt}{2pt}
    \rowcolor{okabe_blue} \textbf{\textit{Real Users}}
            & .887 & .029 & .029 & .945 & .032
            & .931 & .027 & .027 & .953 & .030 \\
    \specialrule{0.9pt}{2pt}{2pt}
    \rowcolor{okabe_pink!20}  \multicolumn{11}{c}{\textit{Closed-Source}} \\
    \midrule
    \texttt{GPT-5.4}
            & 0.857 & 0.307 & 0.282 & 0.141 & 0.302
            & 0.890 & 0.478 & 0.486 & 0.026 & 0.495 \\
    \texttt{GPT-5.4 mini}
            & 0.854 & 0.398 & 0.409 & 0.060 & 0.400
            & 0.893 & 0.507 & 0.494 & 0.020 & 0.520 \\
    \texttt{GPT-5.4 nano}
            & 0.849 & 0.405 & 0.402 & 0.057 & 0.405
            & 0.885 & 0.517 & 0.504 & 0.020 & 0.524 \\
    \texttt{Claude Haiku 4.5}
            & 0.833 & 0.406 & 0.383 & 0.067 & 0.387
            & 0.849 & 0.570 & 0.572 & 0.013 & 0.573 \\
    \texttt{Gemini 3.1 Pro}
            & 0.859 & 0.296 & 0.268 & 0.171 & 0.279
            & 0.885 & 0.488 & 0.482 & 0.026 & 0.494 \\
    \texttt{Gemini 3 Flash}
            & 0.845 & 0.340 & 0.301 & 0.121 & 0.319
            & 0.871 & 0.526 & 0.516 & 0.020 & 0.524 \\
    \texttt{Gemini 3.1 Flash-Lite}
            & 0.824 & 0.426 & 0.398 & 0.055 & 0.410
            & 0.848 & 0.547 & 0.575 & 0.012 & 0.576 \\
    \midrule
    \rowcolor{okabe_blue!20} \multicolumn{11}{c}{\textit{Open-Source}} \\
    \midrule
    \texttt{Qwen3.5-122B-A10B}
            & 0.827 & 0.412 & 0.385 & 0.068 & 0.386
            & 0.861 & 0.533 & 0.560 & 0.015 & 0.553 \\
    \texttt{Qwen3.5-35B-A3B}
            & 0.812 & 0.469 & 0.436 & 0.040 & 0.444
            & 0.843 & 0.567 & 0.597 & 0.010 & 0.594 \\
    \texttt{Qwen3.5-27B}
            & 0.828 & 0.428 & 0.400 & 0.056 & 0.408
            & 0.847 & 0.556 & 0.574 & 0.012 & 0.576 \\
    \texttt{Qwen3.5-9B}
            & 0.839 & 0.412 & 0.382 & 0.060 & 0.400
            & 0.871 & 0.550 & 0.550 & 0.014 & 0.557 \\
    \texttt{Qwen3.5-4B}
            & 0.836 & 0.455 & 0.428 & 0.049 & 0.423
            & 0.867 & 0.577 & 0.581 & 0.012 & 0.574 \\
    \texttt{Qwen3.5-2B}
            & 0.799 & 0.535 & 0.515 & 0.020 & 0.522
            & 0.824 & 0.625 & 0.627 & 0.007 & 0.641 \\
    \texttt{Qwen3.5 0.8B}
            & 0.828 & 0.501 & 0.501 & 0.022 & 0.513
            & 0.856 & 0.586 & 0.590 & 0.009 & 0.605 \\
    \texttt{Llama-3.3-70B-Instruct}
            & 0.835 & 0.434 & 0.408 & 0.053 & 0.414
            & 0.854 & 0.554 & 0.573 & 0.012 & 0.581 \\
    \texttt{Llama-3.1-8B-Instruct}
            & 0.833 & 0.402 & 0.382 & 0.071 & 0.381
            & 0.840 & 0.550 & 0.580 & 0.012 & 0.579 \\
    \texttt{gpt-oss-120b}
            & 0.849 & 0.374 & 0.357 & 0.082 & 0.365
            & 0.885 & 0.492 & 0.508 & 0.020 & 0.519 \\
    \texttt{gpt-oss-20b}
            & 0.848 & 0.500 & 0.389 & 0.066 & 0.389
            & 0.878 & 0.530 & 0.552 & 0.016 & 0.546 \\
    \texttt{gemma-4-31B-it}
            & 0.846 & 0.363 & 0.335 & 0.102 & 0.339
            & 0.866 & 0.524 & 0.548 & 0.017 & 0.540 \\
    \texttt{gemma-4-26B-A4B-it}
            & 0.838 & 0.376 & 0.346 & 0.087 & 0.357
            & 0.854 & 0.570 & 0.571 & 0.013 & 0.567 \\
    \texttt{gemma-4-E4B-it}
            & 0.828 & 0.406 & 0.360 & 0.069 & 0.384
            & 0.849 & 0.569 & 0.567 & 0.014 & 0.565 \\
    \texttt{gemma-4-E2B-it}
            & 0.820 & 0.433 & 0.410 & 0.049 & 0.424
            & 0.847 & 0.566 & 0.581 & 0.011 & 0.588 \\
    \midrule
    \rowcolor{okabe_orange!20}\multicolumn{11}{c}{\textit{Trained Simulators}} \\
    \midrule
    \texttt{UserLM-8b}
            & 0.856 & 0.393 & 0.417 & 0.056 & 0.408
            & 0.900 & 0.428 & 0.438 & 0.039 & 0.449 \\
    \texttt{humanlm-opinion}
            & 0.853 & 0.358 & 0.338 & 0.101 & 0.341
            & 0.886 & 0.491 & 0.508 & 0.023 & 0.505 \\
    \bottomrule
  \end{tabular}}
\end{table*}
\definecolor{okabe_orange}{HTML}{E69F00}
\definecolor{okabe_blue}{HTML}{56B4E9}
\definecolor{okabe_pink}{HTML}{CC79A7}

\begin{table*}[h!]
  \caption{
    Context facet results. The distributional gap between real and simulated user behaviors across coding and writing tasks. We compare the distribution of user behaviors demonstrated in $\mathcal{D}_{\text{real}}$ with those in $\mathcal{D}_{\text{sim}}$. $\uparrow$ indicates higher is better and $\downarrow$ indicates lower is better.
    }
  \label{tab:main_results_context}
  \centering
  \resizebox{\textwidth}{!}{%
  \begin{tabular}{l ccccc ccccc}
    \toprule
    & \multicolumn{5}{c}{\textbf{Coding}} & \multicolumn{5}{c}{\textbf{Writing}} \\
    \cmidrule(lr){2-6} \cmidrule(lr){7-11}
    \textbf{User Simulator}
      & $\mathrm{NN}$\,$\uparrow$
      & $\mathrm{KL_{\mathrm{fwd}}}$\,$\downarrow$
      & $\mathrm{KL_{\mathrm{bwd}}}$\,$\downarrow$
      & $\mathrm{MAU}$\,$\uparrow$
      & $\mathrm{JS}$\,$\downarrow$
      & $\mathrm{NN}$\,$\uparrow$
      & $\mathrm{KL_{\mathrm{fwd}}}$\,$\downarrow$
      & $\mathrm{KL_{\mathrm{bwd}}}$\,$\downarrow$
      & $\mathrm{MAU}$\,$\uparrow$
      & $\mathrm{JS}$\,$\downarrow$ \\
    \specialrule{0.9pt}{2pt}{2pt}
    \rowcolor{okabe_blue} \textbf{\textit{Real Users}}
            & .877 & .025 & .025 & .955 & .029
            & .904 & .020 & .020 & .969 & .023 \\
    \specialrule{0.9pt}{2pt}{2pt}
    \rowcolor{okabe_pink!20}  \multicolumn{11}{c}{\textit{Closed-Source}} \\
    \midrule
    \texttt{GPT-5.4}
            & 0.860 & 0.231 & 0.241 & 0.276 & 0.222
            & 0.879 & 0.341 & 0.358 & 0.099 & 0.343 \\
    \texttt{GPT-5.4 mini}
            & 0.851 & 0.338 & 0.366 & 0.104 & 0.338
            & 0.877 & 0.332 & 0.344 & 0.101 & 0.341 \\
    \texttt{GPT-5.4 nano}
            & 0.850 & 0.305 & 0.313 & 0.152 & 0.293
            & 0.876 & 0.335 & 0.347 & 0.103 & 0.339 \\
    \texttt{Claude Haiku 4.5}
            & 0.841 & 0.378 & 0.356 & 0.089 & 0.356
            & 0.860 & 0.506 & 0.517 & 0.024 & 0.504 \\
    \texttt{Gemini 3.1 Pro}
            & 0.860 & 0.211 & 0.215 & 0.324 & 0.202
            & 0.879 & 0.335 & 0.346 & 0.115 & 0.327 \\
    \texttt{Gemini 3 Flash}
            & 0.845 & 0.343 & 0.324 & 0.117 & 0.325
            & 0.870 & 0.429 & 0.436 & 0.046 & 0.429 \\
    \texttt{Gemini 3.1 Flash-Lite}
            & 0.833 & 0.443 & 0.433 & 0.049 & 0.423
            & 0.860 & 0.514 & 0.521 & 0.022 & 0.513 \\
    \midrule
    \rowcolor{okabe_blue!20} \multicolumn{11}{c}{\textit{Open-Source}} \\
    \midrule
    \texttt{Qwen3.5-122B-A10B}
            & 0.845 & 0.360 & 0.349 & 0.102 & 0.340
            & 0.868 & 0.460 & 0.478 & 0.033 & 0.466 \\
    \texttt{Qwen3.5-35B-A3B}
            & 0.831 & 0.444 & 0.425 & 0.051 & 0.419
            & 0.860 & 0.503 & 0.514 & 0.024 & 0.503 \\
    \texttt{Qwen3.5-27B}
            & 0.832 & 0.425 & 0.407 & 0.058 & 0.404
            & 0.863 & 0.503 & 0.508 & 0.023 & 0.505 \\
    \texttt{Qwen3.5-9B}
            & 0.837 & 0.447 & 0.431 & 0.052 & 0.418
            & 0.864 & 0.498 & 0.503 & 0.024 & 0.499 \\
    \texttt{Qwen3.5-4B}
            & 0.835 & 0.435 & 0.422 & 0.058 & 0.405
            & 0.861 & 0.501 & 0.516 & 0.024 & 0.501 \\
    \texttt{Qwen3.5-2B}
            & 0.832 & 0.468 & 0.470 & 0.034 & 0.464
            & 0.859 & 0.506 & 0.518 & 0.020 & 0.521 \\
    \texttt{Qwen3.5 0.8B}
            & 0.834 & 0.458 & 0.467 & 0.030 & 0.476
            & 0.866 & 0.445 & 0.454 & 0.038 & 0.452 \\
    \texttt{Llama-3.3-70B-Instruct}
            & 0.839 & 0.402 & 0.385 & 0.068 & 0.386
            & 0.858 & 0.498 & 0.510 & 0.023 & 0.506 \\
    \texttt{Llama-3.1-8B-Instruct}
            & 0.838 & 0.388 & 0.376 & 0.077 & 0.373
            & 0.859 & 0.508 & 0.514 & 0.022 & 0.509 \\
    \texttt{gpt-oss-120b}
            & 0.850 & 0.276 & 0.258 & 0.204 & 0.259
            & 0.877 & 0.350 & 0.357 & 0.103 & 0.340 \\
    \texttt{gpt-oss-20b}
            & 0.853 & 0.260 & 0.267 & 0.229 & 0.245
            & 0.876 & 0.355 & 0.367 & 0.094 & 0.349 \\
    \texttt{gemma-4-31B-it}
            & 0.850 & 0.328 & 0.329 & 0.123 & 0.319
            & 0.870 & 0.465 & 0.474 & 0.033 & 0.466 \\
    \texttt{gemma-4-26B-A4B-it}
            & 0.842 & 0.381 & 0.366 & 0.083 & 0.364
            & 0.865 & 0.492 & 0.507 & 0.024 & 0.503 \\
    \texttt{gemma-4-E4B-it}
            & 0.831 & 0.417 & 0.410 & 0.067 & 0.388
            & 0.861 & 0.513 & 0.538 & 0.021 & 0.519 \\
    \texttt{gemma-4-E2B-it}
            & 0.837 & 0.437 & 0.433 & 0.051 & 0.418
            & 0.863 & 0.505 & 0.526 & 0.020 & 0.521 \\
    \midrule
    \rowcolor{okabe_orange!20}\multicolumn{11}{c}{\textit{Trained Simulators}} \\
    \midrule
    \texttt{UserLM-8b}
            & 0.860 & 0.226 & 0.265 & 0.233 & 0.242
            & 0.884 & 0.226 & 0.237 & 0.247 & 0.235 \\
    \texttt{humanlm-opinion}
            & 0.853 & 0.292 & 0.294 & 0.185 & 0.270
            & 0.873 & 0.381 & 0.409 & 0.075 & 0.375 \\
    \bottomrule
  \end{tabular}}
\end{table*}
\definecolor{okabe_orange}{HTML}{E69F00}
\definecolor{okabe_blue}{HTML}{56B4E9}
\definecolor{okabe_pink}{HTML}{CC79A7}

\begin{table*}[h!]
  \caption{
    Communication style facet results. The distributional gap between real and simulated user behaviors across coding and writing tasks. We compare the distribution of user behaviors demonstrated in $\mathcal{D}_{\text{real}}$ with those in $\mathcal{D}_{\text{sim}}$. $\uparrow$ indicates higher is better and $\downarrow$ indicates lower is better.
    }
  \label{tab:main_results_comm_style}
  \centering
  \resizebox{\textwidth}{!}{%
  \begin{tabular}{l ccccc ccccc}
    \toprule
    & \multicolumn{5}{c}{\textbf{Coding}} & \multicolumn{5}{c}{\textbf{Writing}} \\
    \cmidrule(lr){2-6} \cmidrule(lr){7-11}
    \textbf{User Simulator}
      & $\mathrm{NN}$\,$\uparrow$
      & $\mathrm{KL_{\mathrm{fwd}}}$\,$\downarrow$
      & $\mathrm{KL_{\mathrm{bwd}}}$\,$\downarrow$
      & $\mathrm{MAU}$\,$\uparrow$
      & $\mathrm{JS}$\,$\downarrow$
      & $\mathrm{NN}$\,$\uparrow$
      & $\mathrm{KL_{\mathrm{fwd}}}$\,$\downarrow$
      & $\mathrm{KL_{\mathrm{bwd}}}$\,$\downarrow$
      & $\mathrm{MAU}$\,$\uparrow$
      & $\mathrm{JS}$\,$\downarrow$\\
    \specialrule{0.9pt}{2pt}{2pt}
    \rowcolor{okabe_blue} \textbf{\textit{Real Users}}
            & .924 & .026 & .025 & .956 & .028
            & .904 & .020 & .020 & .969 & .023 \\
    \specialrule{0.9pt}{2pt}{2pt}
    \rowcolor{okabe_pink!20}  \multicolumn{11}{c}{\textit{Closed-Source}} \\
    \midrule
    \texttt{GPT-5.4}
            & 0.899 & 0.402 & 0.408 & 0.052 & 0.417
            & 0.891 & 0.494 & 0.504 & 0.021 & 0.517 \\
    \texttt{GPT-5.4 mini}
            & 0.887 & 0.504 & 0.496 & 0.020 & 0.520
            & 0.886 & 0.522 & 0.513 & 0.015 & 0.553 \\
    \texttt{GPT-5.4 nano}
            & 0.886 & 0.507 & 0.498 & 0.019 & 0.530
            & 0.881 & 0.539 & 0.557 & 0.012 & 0.581 \\
    \texttt{Claude Haiku 4.5}
            & 0.832 & 0.607 & 0.612 & 0.006 & 0.656
            & 0.821 & 0.615 & 0.628 & 0.005 & 0.666 \\
    \texttt{Gemini 3.1 Pro}
            & 0.875 & 0.537 & 0.523 & 0.016 & 0.550
            & 0.870 & 0.570 & 0.571 & 0.009 & 0.605 \\
    \texttt{Gemini 3 Flash}
            & 0.861 & 0.570 & 0.574 & 0.010 & 0.596
            & 0.853 & 0.576 & 0.590 & 0.007 & 0.629 \\
    \texttt{Gemini 3.1 Flash-Lite}
            & 0.838 & 0.615 & 0.628 & 0.005 & 0.663
            & 0.833 & 0.609 & 0.610 & 0.006 & 0.661 \\
    \midrule
    \rowcolor{okabe_blue!20} \multicolumn{11}{c}{\textit{Open-Source}} \\
    \midrule
    \texttt{Qwen3.5-122B-A10B}
            & 0.848 & 0.610 & 0.624 & 0.006 & 0.651
            & 0.841 & 0.588 & 0.601 & 0.006 & 0.647 \\
    \texttt{Qwen3.5-35B-A3B}
            & 0.843 & 0.615 & 0.629 & 0.006 & 0.661
            & 0.824 & 0.611 & 0.619 & 0.005 & 0.665 \\
    \texttt{Qwen3.5-27B}
            & 0.851 & 0.598 & 0.611 & 0.006 & 0.647
            & 0.836 & 0.611 & 0.614 & 0.006 & 0.656 \\
    \texttt{Qwen3.5-9B}
            & 0.854 & 0.594 & 0.600 & 0.007 & 0.637
            & 0.857 & 0.584 & 0.596 & 0.007 & 0.637 \\
    \texttt{Qwen3.5-4B}
            & 0.830 & 0.613 & 0.601 & 0.006 & 0.655
            & 0.843 & 0.594 & 0.611 & 0.006 & 0.655 \\
    \texttt{Qwen3.5-2B}
            & 0.801 & 0.622 & 0.624 & 0.005 & 0.682
            & 0.835 & 0.611 & 0.615 & 0.005 & 0.673 \\
    \texttt{Qwen3.5 0.8B}
            & 0.854 & 0.611 & 0.627 & 0.006 & 0.660
            & 0.854 & 0.590 & 0.614 & 0.006 & 0.656 \\
    \texttt{Llama-3.3-70B-Instruct}
            & 0.843 & 0.614 & 0.602 & 0.006 & 0.655
            & 0.841 & 0.597 & 0.616 & 0.005 & 0.663 \\
    \texttt{Llama-3.1-8B-Instruct}
            & 0.858 & 0.573 & 0.575 & 0.008 & 0.615
            & 0.841 & 0.604 & 0.609 & 0.006 & 0.654 \\
    \texttt{gpt-oss-120b}
            & 0.868 & 0.584 & 0.590 & 0.009 & 0.610
            & 0.870 & 0.575 & 0.576 & 0.008 & 0.618 \\
    \texttt{gpt-oss-20b}
            & 0.864 & 0.583 & 0.597 & 0.008 & 0.623
            & 0.862 & 0.589 & 0.599 & 0.007 & 0.638 \\
    \texttt{gemma-4-31B-it}
            & 0.862 & 0.587 & 0.583 & 0.008 & 0.615
            & 0.855 & 0.602 & 0.598 & 0.007 & 0.638 \\
    \texttt{gemma-4-26B-A4B-it}
            & 0.858 & 0.596 & 0.601 & 0.007 & 0.637
            & 0.844 & 0.589 & 0.605 & 0.006 & 0.651 \\
    \texttt{gemma-4-E4B-it}
            & 0.849 & 0.605 & 0.619 & 0.007 & 0.644
            & 0.837 & 0.602 & 0.599 & 0.006 & 0.655 \\
    \texttt{gemma-4-E2B-it}
            & 0.856 & 0.631 & 0.639 & 0.006 & 0.659
            & 0.849 & 0.609 & 0.621 & 0.006 & 0.652 \\
    \midrule
    \rowcolor{okabe_orange!20}\multicolumn{11}{c}{\textit{Trained Simulators}} \\
    \midrule
    \texttt{UserLM-8b}
            & 0.906 & 0.325 & 0.346 & 0.088 & 0.356
            & 0.903 & 0.403 & 0.409 & 0.048 & 0.424 \\
    \texttt{humanlm-opinion}
            & 0.888 & 0.518 & 0.514 & 0.019 & 0.527
            & 0.866 & 0.561 & 0.577 & 0.009 & 0.611 \\
    \bottomrule
  \end{tabular}}
\end{table*}
\definecolor{okabe_orange}{HTML}{E69F00}
\definecolor{okabe_blue}{HTML}{56B4E9}
\definecolor{okabe_pink}{HTML}{CC79A7}

\begin{table*}[h!]
  \caption{
    DAMSL Dialog Acts facet results. The distributional gap between real and simulated user behaviors across coding and writing tasks. We compare the distribution of user behaviors demonstrated in $\mathcal{D}_{\text{real}}$ with those in $\mathcal{D}_{\text{sim}}$. $\uparrow$ indicates higher is better and $\downarrow$ indicates lower is better.
    }
  \label{tab:main_results_damsl}
  \centering
  \resizebox{\textwidth}{!}{%
  \begin{tabular}{l ccccc ccccc}
    \toprule
    & \multicolumn{5}{c}{\textbf{Coding}} & \multicolumn{5}{c}{\textbf{Writing}} \\
    \cmidrule(lr){2-6} \cmidrule(lr){7-11}
    \textbf{User Simulator}
      & $\mathrm{NN}$\,$\uparrow$
      & $\mathrm{KL_{\mathrm{fwd}}}$\,$\downarrow$
      & $\mathrm{KL_{\mathrm{bwd}}}$\,$\downarrow$
      & $\mathrm{MAU}$\,$\uparrow$
      & $\mathrm{JS}$\,$\downarrow$
      & $\mathrm{NN}$\,$\uparrow$
      & $\mathrm{KL_{\mathrm{fwd}}}$\,$\downarrow$
      & $\mathrm{KL_{\mathrm{bwd}}}$\,$\downarrow$
      & $\mathrm{MAU}$\,$\uparrow$
      & $\mathrm{JS}$\,$\downarrow$ \\
    \specialrule{0.9pt}{2pt}{2pt}
    \rowcolor{okabe_blue} \textbf{\textit{Real Users}}
            & .870 & .021 & .022 & .961 & .026
            & .927 & .021 & .021 & .964 & .025 \\
    \specialrule{0.9pt}{2pt}{2pt}
    \rowcolor{okabe_pink!20}  \multicolumn{11}{c}{\textit{Closed-Source}} \\
    \midrule
    \texttt{GPT-5.4}
            & 0.823 & 0.432 & 0.451 & 0.046 & 0.429
            & 0.866 & 0.555 & 0.556 & 0.013 & 0.568 \\
    \texttt{GPT-5.4 mini}
            & 0.814 & 0.512 & 0.539 & 0.019 & 0.526
            & 0.862 & 0.569 & 0.573 & 0.010 & 0.598 \\
    \texttt{GPT-5.4 nano}
            & 0.810 & 0.527 & 0.553 & 0.018 & 0.536
            & 0.864 & 0.562 & 0.571 & 0.010 & 0.597 \\
    \texttt{Claude Haiku 4.5}
            & 0.813 & 0.428 & 0.406 & 0.057 & 0.406
            & 0.852 & 0.576 & 0.563 & 0.014 & 0.565 \\
    \texttt{Gemini 3.1 Pro}
            & 0.827 & 0.428 & 0.432 & 0.048 & 0.425
            & 0.871 & 0.521 & 0.533 & 0.015 & 0.551 \\
    \texttt{Gemini 3 Flash}
            & 0.817 & 0.423 & 0.404 & 0.061 & 0.398
            & 0.868 & 0.526 & 0.547 & 0.017 & 0.541 \\
    \texttt{Gemini 3.1 Flash-Lite}
            & 0.809 & 0.452 & 0.418 & 0.049 & 0.423
            & 0.863 & 0.546 & 0.548 & 0.016 & 0.544 \\
    \midrule
    \rowcolor{okabe_blue!20} \multicolumn{11}{c}{\textit{Open-Source}} \\
    \midrule
    \texttt{Qwen3.5-122B-A10B}
            & 0.815 & 0.428 & 0.403 & 0.060 & 0.401
            & 0.863 & 0.534 & 0.542 & 0.017 & 0.540 \\
    \texttt{Qwen3.5-35B-A3B}
            & 0.801 & 0.470 & 0.455 & 0.038 & 0.452
            & 0.856 & 0.583 & 0.586 & 0.011 & 0.589 \\
    \texttt{Qwen3.5-27B}
            & 0.803 & 0.467 & 0.437 & 0.042 & 0.440
            & 0.861 & 0.572 & 0.570 & 0.013 & 0.573 \\
    \texttt{Qwen3.5-9B}
            & 0.800 & 0.486 & 0.460 & 0.033 & 0.466
            & 0.859 & 0.567 & 0.566 & 0.013 & 0.572 \\
    \texttt{Qwen3.5-4B}
            & 0.803 & 0.502 & 0.482 & 0.030 & 0.479
            & 0.854 & 0.583 & 0.589 & 0.010 & 0.594 \\
    \texttt{Qwen3.5-2B}
            & 0.780 & 0.617 & 0.628 & 0.008 & 0.625
            & 0.831 & 0.638 & 0.622 & 0.006 & 0.654 \\
    \texttt{Qwen3.5 0.8B}
            & 0.781 & 0.608 & 0.629 & 0.007 & 0.636
            & 0.834 & 0.636 & 0.641 & 0.006 & 0.655 \\
    \texttt{Llama-3.3-70B-Instruct}
            & 0.808 & 0.445 & 0.424 & 0.052 & 0.416
            & 0.849 & 0.571 & 0.564 & 0.012 & 0.579 \\
    \texttt{Llama-3.1-8B-Instruct}
            & 0.808 & 0.430 & 0.412 & 0.054 & 0.412
            & 0.849 & 0.576 & 0.578 & 0.012 & 0.579 \\
    \texttt{gpt-oss-120b}
            & 0.817 & 0.436 & 0.427 & 0.050 & 0.421
            & 0.868 & 0.529 & 0.541 & 0.015 & 0.552 \\
    \texttt{gpt-oss-20b}
            & 0.809 & 0.517 & 0.538 & 0.021 & 0.519
            & 0.858 & 0.580 & 0.599 & 0.009 & 0.606 \\
    \texttt{gemma-4-31B-it}
            & 0.822 & 0.390 & 0.368 & 0.078 & 0.370
            & 0.866 & 0.536 & 0.536 & 0.018 & 0.535 \\
    \texttt{gemma-4-26B-A4B-it}
            & 0.815 & 0.449 & 0.442 & 0.046 & 0.431
            & 0.861 & 0.582 & 0.589 & 0.011 & 0.591 \\
    \texttt{gemma-4-E4B-it}
            & 0.810 & 0.411 & 0.368 & 0.069 & 0.384
            & 0.862 & 0.525 & 0.521 & 0.020 & 0.525 \\
    \texttt{gemma-4-E2B-it}
            & 0.812 & 0.422 & 0.392 & 0.058 & 0.405
            & 0.862 & 0.534 & 0.536 & 0.015 & 0.552 \\
    \midrule
    \rowcolor{okabe_orange!20}\multicolumn{11}{c}{\textit{Trained Simulators}} \\
    \midrule
    \texttt{UserLM-8b}
            & 0.835 & 0.398 & 0.439 & 0.046 & 0.431
            & 0.891 & 0.417 & 0.436 & 0.045 & 0.432 \\
    \texttt{humanlm-opinion}
            & 0.821 & 0.433 & 0.437 & 0.046 & 0.430
            & 0.863 & 0.560 & 0.578 & 0.012 & 0.579 \\
    \bottomrule
  \end{tabular}}
\end{table*}
\definecolor{okabe_orange}{HTML}{E69F00}
\definecolor{okabe_blue}{HTML}{56B4E9}
\definecolor{okabe_pink}{HTML}{CC79A7}

\begin{table*}[h!]
  \caption{
    SGD Dialog Acts facet results. The distributional gap between real and simulated user behaviors across coding and writing tasks. We compare the distribution of user behaviors demonstrated in $\mathcal{D}_{\text{real}}$ with those in $\mathcal{D}_{\text{sim}}$. $\uparrow$ indicates higher is better and $\downarrow$ indicates lower is better.
    }
  \label{tab:main_results_sgd}
  \centering
  \resizebox{\textwidth}{!}{%
  \begin{tabular}{l ccccc ccccc}
    \toprule
    & \multicolumn{5}{c}{\textbf{Coding}} & \multicolumn{5}{c}{\textbf{Writing}} \\
    \cmidrule(lr){2-6} \cmidrule(lr){7-11}
    \textbf{User Simulator}
      & $\mathrm{NN}$\,$\uparrow$
      & $\mathrm{KL_{\mathrm{fwd}}}$\,$\downarrow$
      & $\mathrm{KL_{\mathrm{bwd}}}$\,$\downarrow$
      & $\mathrm{MAU}$\,$\uparrow$
      & $\mathrm{JS}$\,$\downarrow$
      & $\mathrm{NN}$\,$\uparrow$
      & $\mathrm{KL_{\mathrm{fwd}}}$\,$\downarrow$
      & $\mathrm{KL_{\mathrm{bwd}}}$\,$\downarrow$
      & $\mathrm{MAU}$\,$\uparrow$
      & $\mathrm{JS}$\,$\downarrow$ \\
    \specialrule{0.9pt}{2pt}{2pt}
    \rowcolor{okabe_blue} \textbf{\textit{Real Users}}
            & .993 & .630 & .414 & .117 & .321
            & .997 & .490 & .501 & .108 & .331 \\
    \specialrule{0.9pt}{2pt}{2pt}
    \rowcolor{okabe_pink!20}  \multicolumn{11}{c}{\textit{Closed-Source}} \\
    \midrule
    \texttt{GPT-5.4}
            & 0.941 & 0.948 & 0.615 & 0.008 & 0.627
            & 0.943 & 0.792 & 0.638 & 0.005 & 0.664 \\
    \texttt{GPT-5.4 mini}
            & 0.937 & 0.987 & 0.737 & 0.006 & 0.657
            & 0.964 & 0.626 & 0.722 & 0.005 & 0.669 \\
    \texttt{GPT-5.4 nano}
            & 0.935 & 0.973 & 0.637 & 0.006 & 0.660
            & 0.934 & 0.804 & 0.619 & 0.005 & 0.674 \\
    \texttt{Claude Haiku 4.5}
            & 0.909 & 0.964 & 0.610 & 0.006 & 0.661
            & 0.880 & 0.837 & 0.608 & 0.005 & 0.677 \\
    \texttt{Gemini 3.1 Pro}
            & 0.902 & 0.990 & 0.696 & 0.005 & 0.676
            & 0.890 & 0.814 & 0.647 & 0.005 & 0.684 \\
    \texttt{Gemini 3 Flash}
            & 0.924 & 0.945 & 0.602 & 0.006 & 0.650
            & 0.950 & 0.800 & 0.601 & 0.005 & 0.667 \\
    \texttt{Gemini 3.1 Flash-Lite}
            & 0.896 & 0.985 & 0.644 & 0.005 & 0.674
            & 0.902 & 0.800 & 0.611 & 0.005 & 0.675 \\
    \midrule
    \rowcolor{okabe_blue!20} \multicolumn{11}{c}{\textit{Open-Source}} \\
    \midrule
    \texttt{Qwen3.5-122B-A10B}
            & 0.900 & 0.970 & 0.630 & 0.005 & 0.672
            & 0.899 & 0.864 & 0.613 & 0.005 & 0.676 \\
    \texttt{Qwen3.5-35B-A3B}
            & 0.906 & 0.968 & 0.621 & 0.005 & 0.669
            & 0.891 & 0.828 & 0.610 & 0.005 & 0.681 \\
    \texttt{Qwen3.5-27B}
            & 0.915 & 0.971 & 0.626 & 0.005 & 0.670
            & 0.897 & 0.821 & 0.612 & 0.005 & 0.679 \\
    \texttt{Qwen3.5-9B}
            & 0.913 & 0.952 & 0.591 & 0.006 & 0.648
            & 0.911 & 0.791 & 0.594 & 0.005 & 0.667 \\
    \texttt{Qwen3.5-4B}
            & 0.906 & 0.963 & 0.605 & 0.005 & 0.666
            & 0.895 & 0.834 & 0.604 & 0.005 & 0.677 \\
    \texttt{Qwen3.5-2B}
            & 0.901 & 0.962 & 0.606 & 0.005 & 0.666
            & 0.888 & 0.845 & 0.601 & 0.005 & 0.680 \\
    \texttt{Qwen3.5 0.8B}
            & 0.939 & 0.948 & 0.596 & 0.006 & 0.647
            & 0.934 & 0.810 & 0.595 & 0.005 & 0.668 \\
    \texttt{Llama-3.3-70B-Instruct}
            & 0.914 & 0.965 & 0.611 & 0.005 & 0.669
            & 0.907 & 0.829 & 0.616 & 0.005 & 0.676 \\
    \texttt{Llama-3.1-8B-Instruct}
            & 0.918 & 0.958 & 0.597 & 0.006 & 0.657
            & 0.905 & 0.799 & 0.606 & 0.005 & 0.672 \\
    \texttt{gpt-oss-120b}
            & 0.954 & 0.939 & 0.596 & 0.008 & 0.621
            & 0.960 & 0.775 & 0.597 & 0.006 & 0.653 \\
    \texttt{gpt-oss-20b}
            & 0.942 & 0.960 & 0.605 & 0.006 & 0.655
            & 0.971 & 0.807 & 0.597 & 0.005 & 0.665 \\
    \texttt{gemma-4-31B-it}
            & 0.893 & 0.989 & 0.632 & 0.005 & 0.670
            & 0.887 & 0.831 & 0.607 & 0.005 & 0.680 \\
    \texttt{gemma-4-26B-A4B-it}
            & 0.923 & 0.957 & 0.608 & 0.006 & 0.658
            & 0.898 & 0.828 & 0.604 & 0.005 & 0.676 \\
    \texttt{gemma-4-E4B-it}
            & 0.939 & 0.946 & 0.589 & 0.007 & 0.635
            & 0.956 & 0.816 & 0.593 & 0.006 & 0.659 \\
    \texttt{gemma-4-E2B-it}
            & 0.921 & 0.965 & 0.614 & 0.006 & 0.658
            & 0.932 & 0.796 & 0.602 & 0.005 & 0.671 \\
    \midrule
    \rowcolor{okabe_orange!20}\multicolumn{11}{c}{\textit{Trained Simulators}} \\
    \midrule
    \texttt{UserLM-8b}
            & 0.986 & 0.802 & 0.517 & 0.023 & 0.505
            & 0.992 & 0.504 & 0.430 & 0.037 & 0.454 \\
    \texttt{humanlm-opinion}
            & 0.964 & 0.905 & 0.595 & 0.011 & 0.591
            & 0.987 & 0.648 & 0.525 & 0.015 & 0.554 \\
    \bottomrule
  \end{tabular}}
\end{table*}

\clearpage
\newpage
\section{Human Study Details}
\label{appendix:humman_study}

\begin{figure*}[h]
    \centering
    \includegraphics[width=\textwidth]{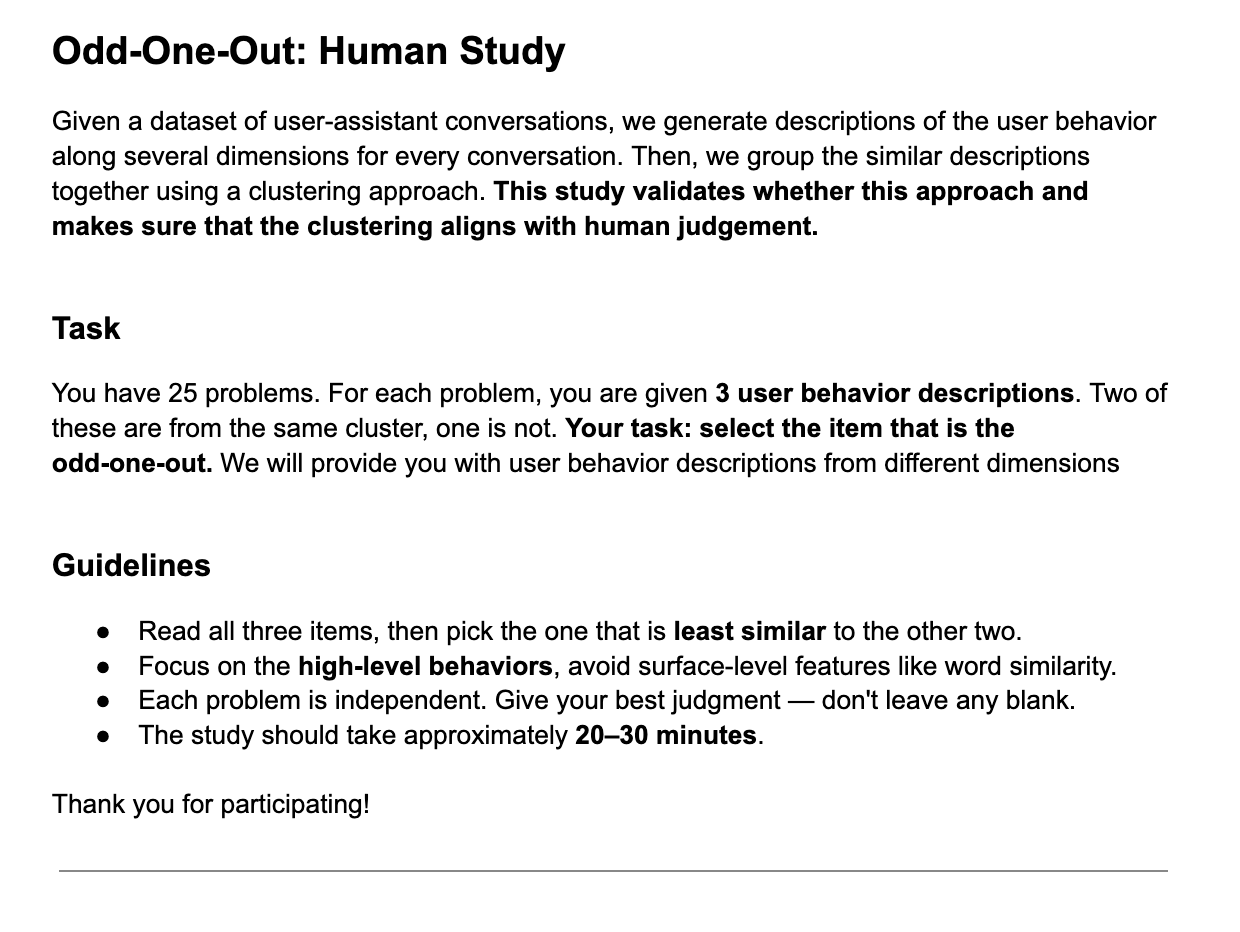}
    \caption{Instructions provided to human annotators for the "Odd-One-Out" human study.}
    \label{fig:human_study}
\end{figure*}

15 annotators were recruited for the human study. The annotators were graduate-level computer science students with a background in natural language processing, recruiting through the university as volunteers. A screenshot of the instructions given to annotators is provided in Figure \ref{fig:human_study}.

\clearpage
\newpage
\section{Ablation Results}
\label{appendix:ablation_results}

\definecolor{okabe_orange}{HTML}{E69F00}
\definecolor{okabe_blue}{HTML}{56B4E9}
\definecolor{okabe_pink}{HTML}{CC79A7}

\begin{table*}[h!]
  \caption{Ablation experiments for user behavior representation: the distributional gap between real and simulated user behaviors across coding and writing tasks when using the raw conversations, usser utterances only, and our behavioral descriptions. We report nearest neighbor cosine similarity ($\mathrm{NN}$), forward KL divergence ($\mathrm{KL_{fwd}}$), backward KL divergence ($\mathrm{KL_{bwd}}$), MAUVE ($\mathrm{MAU}$), and Jensen--Shannon divergence ($\mathrm{JS}$). $\uparrow$ indicates higher is better and $\downarrow$ indicates lower is better.}
  \label{tab:representation_ablation}
  \centering
  \tiny
  \resizebox{\textwidth}{!}{%
  \begin{tabular}{ll ccccc ccccc}
    \toprule
    & & \multicolumn{5}{c}{\textbf{Coding}} & \multicolumn{5}{c}{\textbf{Writing}} \\
    \cmidrule(lr){3-7} \cmidrule(lr){8-12}
    \textbf{User Simulator} & \textbf{Representation}
      & $\mathrm{NN}$\,$\uparrow$
      & $\mathrm{KL_{\mathrm{fwd}}}$\,$\downarrow$
      & $\mathrm{KL_{\mathrm{bwd}}}$\,$\downarrow$
      & $\mathrm{MAU}$\,$\uparrow$
      & $\mathrm{JS}$\,$\downarrow$
      & $\mathrm{NN}$\,$\uparrow$
      & $\mathrm{KL_{\mathrm{fwd}}}$\,$\downarrow$
      & $\mathrm{KL_{\mathrm{bwd}}}$\,$\downarrow$
      & $\mathrm{MAU}$\,$\uparrow$
      & $\mathrm{JS}$\,$\downarrow$ \\
    \midrule
    \rowcolor{okabe_pink!20}  \multicolumn{12}{c}{\textit{Closed-Source}} \\
    \midrule
    \texttt{GPT-5.4}
      & Raw Conversation
            & 0.742 & 0.056 & 0.060 & 0.862 & 0.054
            & 0.737 & 0.300 & 0.305 & 0.165 & 0.282 \\
      & Raw User Utterances
            & 0.721 & 0.071 & 0.077 & 0.798 & 0.070
            & 0.724 & 0.267 & 0.277 & 0.219 & 0.249 \\
      & Behavioral Descriptions
            & 0.872 & 0.261 & 0.265 & 0.223 & 0.248
            & 0.900 & 0.482 & 0.489 & 0.030 & 0.479 \\
    \midrule
    \texttt{GPT-5.4 mini}
      & Raw Conversation
            & 0.744 & 0.067 & 0.075 & 0.833 & 0.062
            & 0.733 & 0.293 & 0.311 & 0.156 & 0.289 \\
      & Raw User Utterances
            & 0.725 & 0.097 & 0.101 & 0.714 & 0.090
            & 0.724 & 0.277 & 0.303 & 0.199 & 0.259 \\
      & Behavioral Descriptions
            & 0.864 & 0.363 & 0.388 & 0.082 & 0.364
            & 0.897 & 0.513 & 0.533 & 0.020 & 0.520 \\
    \midrule
    \texttt{GPT-5.4 nano}
      & Raw Conversation
            & 0.727 & 0.074 & 0.081 & 0.778 & 0.074
            & 0.714 & 0.360 & 0.395 & 0.085 & 0.359 \\
      & Raw User Utterances
            & 0.715 & 0.079 & 0.082 & 0.772 & 0.076
            & 0.708 & 0.296 & 0.319 & 0.159 & 0.287 \\
      & Behavioral Descriptions
            & 0.860 & 0.353 & 0.364 & 0.091 & 0.353
            & 0.896 & 0.520 & 0.531 & 0.018 & 0.532 \\
    \midrule
    \texttt{Claude Haiku 4.5}
      & Raw Conversation
            & 0.733 & 0.044 & 0.041 & 0.910 & 0.042
            & 0.712 & 0.357 & 0.371 & 0.096 & 0.345 \\
      & Raw User Utterances
            & 0.709 & 0.072 & 0.072 & 0.793 & 0.071
            & 0.698 & 0.353 & 0.372 & 0.105 & 0.335 \\
      & Behavioral Descriptions
            & 0.851 & 0.438 & 0.430 & 0.048 & 0.427
            & 0.873 & 0.605 & 0.608 & 0.009 & 0.613 \\
    \midrule
    \texttt{Gemini 3.1 Pro}
      & Raw Conversation
            & 0.747 & 0.059 & 0.058 & 0.843 & 0.059
            & 0.734 & 0.323 & 0.334 & 0.120 & 0.319 \\
      & Raw User Utterances
            & 0.712 & 0.120 & 0.126 & 0.635 & 0.109
            & 0.710 & 0.317 & 0.329 & 0.155 & 0.290 \\
      & Behavioral Descriptions
            & 0.870 & 0.260 & 0.256 & 0.226 & 0.246
            & 0.899 & 0.482 & 0.471 & 0.033 & 0.466 \\
    \midrule
    \texttt{Gemini 3 Flash}
      & Raw Conversation
            & 0.730 & 0.051 & 0.049 & 0.868 & 0.053
            & 0.714 & 0.340 & 0.356 & 0.106 & 0.334 \\
      & Raw User Utterances
            & 0.701 & 0.089 & 0.083 & 0.754 & 0.080
            & 0.706 & 0.259 & 0.282 & 0.219 & 0.248 \\
      & Behavioral Descriptions
            & 0.858 & 0.355 & 0.328 & 0.108 & 0.334
            & 0.888 & 0.539 & 0.545 & 0.018 & 0.533 \\
    \midrule
    \texttt{Gemini 3.1 Flash-Lite}
      & Raw Conversation
            & 0.720 & 0.054 & 0.048 & 0.865 & 0.053
            & 0.708 & 0.352 & 0.356 & 0.108 & 0.332 \\
      & Raw User Utterances
            & 0.695 & 0.093 & 0.087 & 0.722 & 0.088
            & 0.690 & 0.339 & 0.368 & 0.109 & 0.330 \\
      & Behavioral Descriptions
            & 0.847 & 0.432 & 0.406 & 0.050 & 0.421
            & 0.881 & 0.583 & 0.586 & 0.011 & 0.582 \\
    \midrule
    \rowcolor{okabe_blue!20} \multicolumn{12}{c}{\textit{Open-Source}} \\
    \midrule
    \texttt{Qwen3.5-122B-A10B}
      & Raw Conversation
            & 0.734 & 0.043 & 0.041 & 0.920 & 0.039
            & 0.717 & 0.360 & 0.374 & 0.098 & 0.343 \\
      & Raw User Utterances
            & 0.710 & 0.076 & 0.074 & 0.782 & 0.074
            & 0.707 & 0.321 & 0.333 & 0.149 & 0.294 \\
      & Behavioral Descriptions
            & 0.854 & 0.394 & 0.378 & 0.068 & 0.387
            & 0.883 & 0.581 & 0.575 & 0.012 & 0.579 \\
    \midrule
    \texttt{Qwen3.5-35B-A3B}
      & Raw Conversation
            & 0.724 & 0.052 & 0.050 & 0.870 & 0.053
            & 0.715 & 0.334 & 0.348 & 0.107 & 0.332 \\
      & Raw User Utterances
            & 0.702 & 0.088 & 0.082 & 0.741 & 0.084
            & 0.703 & 0.324 & 0.333 & 0.124 & 0.316 \\
      & Behavioral Descriptions
            & 0.843 & 0.480 & 0.467 & 0.030 & 0.478
            & 0.876 & 0.602 & 0.609 & 0.009 & 0.615 \\
    \midrule
    \texttt{Qwen3.5-27B}
      & Raw Conversation
            & 0.729 & 0.051 & 0.050 & 0.878 & 0.051
            & 0.720 & 0.348 & 0.366 & 0.109 & 0.330 \\
      & Raw User Utterances
            & 0.703 & 0.087 & 0.084 & 0.750 & 0.081
            & 0.705 & 0.328 & 0.356 & 0.125 & 0.315 \\
      & Behavioral Descriptions
            & 0.847 & 0.437 & 0.421 & 0.045 & 0.433
            & 0.881 & 0.570 & 0.580 & 0.011 & 0.584 \\
    \midrule
    \texttt{Qwen3.5-9B}
      & Raw Conversation
            & 0.719 & 0.060 & 0.058 & 0.853 & 0.057
            & 0.706 & 0.331 & 0.344 & 0.112 & 0.327 \\
      & Raw User Utterances
            & 0.694 & 0.101 & 0.098 & 0.703 & 0.093
            & 0.695 & 0.306 & 0.319 & 0.163 & 0.284 \\
      & Behavioral Descriptions
            & 0.850 & 0.465 & 0.457 & 0.037 & 0.454
            & 0.882 & 0.571 & 0.575 & 0.011 & 0.582 \\
    \midrule
    \texttt{Qwen3.5-4B}
      & Raw Conversation
            & 0.716 & 0.059 & 0.056 & 0.859 & 0.056
            & 0.701 & 0.355 & 0.360 & 0.097 & 0.344 \\
      & Raw User Utterances
            & 0.688 & 0.089 & 0.086 & 0.743 & 0.083
            & 0.692 & 0.301 & 0.309 & 0.163 & 0.284 \\
      & Behavioral Descriptions
            & 0.848 & 0.482 & 0.471 & 0.032 & 0.472
            & 0.878 & 0.597 & 0.609 & 0.009 & 0.609 \\
    \midrule
    \texttt{Qwen3.5-2B}
      & Raw Conversation
            & 0.680 & 0.146 & 0.161 & 0.478 & 0.149
            & 0.667 & 0.431 & 0.446 & 0.052 & 0.414 \\
      & Raw User Utterances
            & 0.673 & 0.158 & 0.162 & 0.483 & 0.148
            & 0.666 & 0.362 & 0.381 & 0.090 & 0.353 \\
      & Behavioral Descriptions
            & 0.824 & 0.584 & 0.591 & 0.009 & 0.608
            & 0.858 & 0.630 & 0.642 & 0.006 & 0.660 \\
    \midrule
    \texttt{Qwen3.5 0.8B}
      & Raw Conversation
            & 0.691 & 0.152 & 0.183 & 0.463 & 0.154
            & 0.671 & 0.426 & 0.455 & 0.046 & 0.429 \\
      & Raw User Utterances
            & 0.685 & 0.132 & 0.146 & 0.533 & 0.134
            & 0.667 & 0.374 & 0.382 & 0.086 & 0.359 \\
      & Behavioral Descriptions
            & 0.832 & 0.547 & 0.568 & 0.012 & 0.579
            & 0.869 & 0.629 & 0.620 & 0.006 & 0.648 \\
    \midrule
    \texttt{Llama-3.3-70B-Instruct}
      & Raw Conversation
            & 0.730 & 0.043 & 0.041 & 0.907 & 0.043
            & 0.708 & 0.347 & 0.338 & 0.105 & 0.335 \\
      & Raw User Utterances
            & 0.696 & 0.095 & 0.086 & 0.735 & 0.085
            & 0.688 & 0.319 & 0.314 & 0.147 & 0.296 \\
      & Behavioral Descriptions
            & 0.848 & 0.471 & 0.456 & 0.033 & 0.469
            & 0.870 & 0.601 & 0.602 & 0.008 & 0.615 \\
    \midrule
    \texttt{Llama-3.1-8B-Instruct}
      & Raw Conversation
            & 0.708 & 0.065 & 0.063 & 0.828 & 0.063
            & 0.693 & 0.346 & 0.345 & 0.099 & 0.341 \\
      & Raw User Utterances
            & 0.674 & 0.096 & 0.088 & 0.713 & 0.090
            & 0.671 & 0.340 & 0.344 & 0.121 & 0.319 \\
      & Behavioral Descriptions
            & 0.849 & 0.435 & 0.427 & 0.049 & 0.423
            & 0.870 & 0.598 & 0.610 & 0.009 & 0.608 \\
    \midrule
    \texttt{gpt-oss-120b}
      & Raw Conversation
            & 0.711 & 0.065 & 0.067 & 0.827 & 0.063
            & 0.710 & 0.343 & 0.360 & 0.104 & 0.336 \\
      & Raw User Utterances
            & 0.695 & 0.075 & 0.075 & 0.779 & 0.075
            & 0.699 & 0.312 & 0.315 & 0.149 & 0.294 \\
      & Behavioral Descriptions
            & 0.860 & 0.358 & 0.353 & 0.100 & 0.343
            & 0.894 & 0.532 & 0.549 & 0.017 & 0.537 \\
    \midrule
    \texttt{gpt-oss-20b}
      & Raw Conversation
            & 0.704 & 0.091 & 0.095 & 0.708 & 0.091
            & 0.682 & 0.381 & 0.408 & 0.066 & 0.389 \\
      & Raw User Utterances
            & 0.702 & 0.083 & 0.083 & 0.747 & 0.082
            & 0.679 & 0.363 & 0.369 & 0.098 & 0.343 \\
      & Behavioral Descriptions
            & 0.857 & 0.411 & 0.435 & 0.055 & 0.409
            & 0.888 & 0.569 & 0.593 & 0.012 & 0.575 \\
    \midrule
    \texttt{gemma-4-31B-it}
      & Raw Conversation
            & 0.740 & 0.046 & 0.043 & 0.901 & 0.044
            & 0.723 & 0.341 & 0.359 & 0.103 & 0.337 \\
      & Raw User Utterances
            & 0.708 & 0.098 & 0.099 & 0.711 & 0.091
            & 0.703 & 0.323 & 0.342 & 0.127 & 0.313 \\
      & Behavioral Descriptions
            & 0.862 & 0.311 & 0.296 & 0.148 & 0.297
            & 0.889 & 0.537 & 0.547 & 0.016 & 0.545 \\
    \midrule
    \texttt{gemma-4-26B-A4B-it}
      & Raw Conversation
            & 0.728 & 0.055 & 0.055 & 0.867 & 0.053
            & 0.720 & 0.353 & 0.348 & 0.107 & 0.333 \\
      & Raw User Utterances
            & 0.689 & 0.122 & 0.120 & 0.630 & 0.110
            & 0.696 & 0.341 & 0.351 & 0.122 & 0.317 \\
      & Behavioral Descriptions
            & 0.856 & 0.397 & 0.373 & 0.073 & 0.379
            & 0.882 & 0.577 & 0.585 & 0.011 & 0.587 \\
    \midrule
    \texttt{gemma-4-E4B-it}
      & Raw Conversation
            & 0.716 & 0.052 & 0.050 & 0.884 & 0.049
            & 0.704 & 0.328 & 0.336 & 0.122 & 0.317 \\
      & Raw User Utterances
            & 0.686 & 0.073 & 0.067 & 0.808 & 0.068
            & 0.682 & 0.335 & 0.336 & 0.123 & 0.317 \\
      & Behavioral Descriptions
            & 0.846 & 0.422 & 0.383 & 0.059 & 0.402
            & 0.876 & 0.589 & 0.591 & 0.010 & 0.597 \\
    \midrule
    \texttt{gemma-4-E2B-it}
      & Raw Conversation
            & 0.726 & 0.052 & 0.049 & 0.878 & 0.051
            & 0.709 & 0.315 & 0.323 & 0.140 & 0.302 \\
      & Raw User Utterances
            & 0.683 & 0.109 & 0.101 & 0.678 & 0.098
            & 0.676 & 0.334 & 0.328 & 0.121 & 0.319 \\
      & Behavioral Descriptions
            & 0.849 & 0.446 & 0.435 & 0.044 & 0.435
            & 0.879 & 0.586 & 0.589 & 0.010 & 0.599 \\
    \midrule
    \rowcolor{okabe_orange!20}\multicolumn{12}{c}{\textit{Trained Simulators}} \\
    \midrule
    \texttt{UserLM-8b}
      & Raw Conversation
            & 0.695 & 0.152 & 0.210 & 0.406 & 0.170
            & 0.695 & 0.356 & 0.426 & 0.085 & 0.361 \\
      & Raw User Utterances
            & 0.692 & 0.153 & 0.213 & 0.438 & 0.160
            & 0.688 & 0.300 & 0.365 & 0.148 & 0.295 \\
      & Behavioral Descriptions
            & 0.867 & 0.328 & 0.388 & 0.093 & 0.349
            & 0.905 & 0.392 & 0.424 & 0.054 & 0.411 \\
    \midrule
    \texttt{humanlm-opinion}
      & Raw Conversation
            & 0.728 & 0.049 & 0.047 & 0.895 & 0.046
            & 0.717 & 0.284 & 0.301 & 0.172 & 0.278 \\
      & Raw User Utterances
            & 0.706 & 0.077 & 0.075 & 0.786 & 0.073
            & 0.701 & 0.279 & 0.283 & 0.204 & 0.257 \\
      & Behavioral Descriptions
            & 0.867 & 0.268 & 0.257 & 0.225 & 0.247
            & 0.892 & 0.514 & 0.519 & 0.018 & 0.531 \\
    \bottomrule
  \end{tabular}}
\end{table*}

\begin{figure*}[h]
    \centering
    \includegraphics[width=0.8\textwidth]{figures/ablations.jpg}
    \caption{Scatter plot visualization of Spearman rank correlation ($\rho$) of simulator rankings across different embedding models and clustering algorithms. Each point represents a simulator, with axes showing its z-score normalized metric values under two variants. Points near the diagonal and high $\rho$ values indicate that the relative ordering of simulators is preserved across variants. each dataset. (a) compares three embedding models: \texttt{Qwen3-Embedding-8B} (Qwen3) \citep{qwen3embedding}, texttt{e5-large-v2} (E5) \citep{wang2022text} and \texttt{BGE-small-en-v1.5} (BGE) \citep{bge_embedding}. (b) compares three clustering algorithms:  $k$-means (KM), Gaussian Mixture Models (GMM) and Agglomerative Clustering (Agg) ($k{=}500$).}
    \label{fig:ablations}
\end{figure*}

\clearpage
\newpage
\section{Linear Classification Results}
\label{appendix:linear_classification}
\begin{table}[h!]
  \caption{
    Classification accuracy of an L2-regularized logistic regression classifier trained to distinguish real from simulated user behavior representation embeddings for each simulator. Results are averaged over 5 stratified random 80/20 train-test splits.}
  \label{tab:linear_classification}
  \centering
  \footnotesize
  \begin{tabular}{lcc}
    \toprule
    \textbf{User Simulator} & \textbf{Coding} & \textbf{Writing} \\
    \midrule
    \rowcolor{okabe_pink!20} \multicolumn{3}{c}{\textit{Closed-Source}} \\
    \midrule
    \texttt{GPT-5.4} & 0.9466 & 0.9677 \\
    \texttt{GPT-5.4 mini} & 0.9662 & 0.9821 \\
    \texttt{GPT-5.4 nano} & 0.9701 & 0.9810 \\
    \texttt{Claude Haiku 4.5} & 0.9825 & 0.9909 \\
    \texttt{Gemini 3.1 Pro} & 0.9639 & 0.9717 \\
    \texttt{Gemini 3 Flash} & 0.9671 & 0.9766 \\
    \texttt{Gemini 3.1 Flash-Lite} & 0.9890 & 0.9879 \\
    \midrule
    \rowcolor{okabe_blue!20} \multicolumn{3}{c}{\textit{Open-Source}} \\
    \midrule
    \texttt{Qwen3.5-122B-A10B} & 0.9859 & 0.9873 \\
    \texttt{Qwen3.5-35B-A3B} & 0.9867 & 0.9908 \\
    \texttt{Qwen3.5-27B} & 0.9833 & 0.9903 \\
    \texttt{Qwen3.5-9B} & 0.9761 & 0.9847 \\
    \texttt{Qwen3.5-4B} & 0.9839 & 0.9874 \\
    \texttt{Qwen3.5-2B} & 0.9930 & 0.9963 \\
    \texttt{Qwen3.5 0.8B} & 0.9888 & 0.9946 \\
    \texttt{Llama-3.3-70B-Instruct} & 0.9837 & 0.9923 \\
    \texttt{Llama-3.1-8B-Instruct} & 0.9726 & 0.9895 \\
    \texttt{gpt-oss-120b} & 0.9701 & 0.9768 \\
    \texttt{gpt-oss-20b} & 0.9781 & 0.9854 \\
    \texttt{gemma-4-31B-it} & 0.9747 & 0.9790 \\
    \texttt{gemma-4-26B-A4B-it} & 0.9792 & 0.9869 \\
    \texttt{gemma-4-E4B-it} & 0.9793 & 0.9859 \\
    \texttt{gemma-4-E2B-it} & 0.9839 & 0.9895 \\
    \midrule
    \rowcolor{okabe_orange!20} \multicolumn{3}{c}{\textit{Trained Simulators}} \\
    \midrule
    \texttt{UserLM-8b} & 0.9092 & 0.9266 \\
    \texttt{humanlm-opinion} & 0.9660 & 0.9763 \\
    \bottomrule
  \end{tabular}
\end{table}

\begin{figure*}[h]
    \centering
    \includegraphics[width=\textwidth]{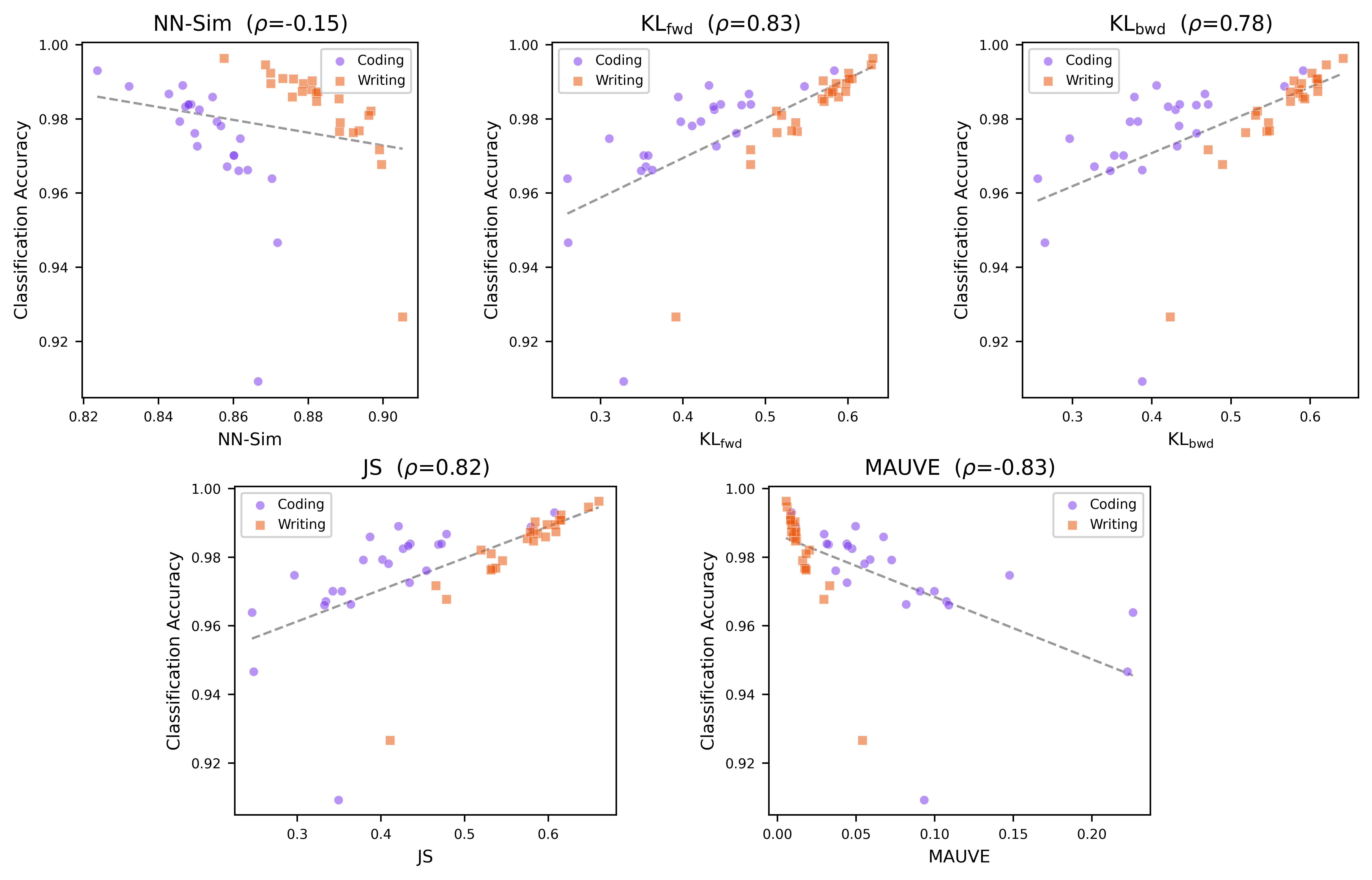}
    \caption{Scatter plots of classification accuracy and each distributional metric for all simulators across coding and writing tasks: nearest neighbor cosine similarity ($\mathrm{NN-Sim}$), forward KL divergence ($\mathrm{KL_{fwd}}$), backward KL divergence ($\mathrm{KL_{bwd}}$), Jensen--Shannon divergence ($\mathrm{JS}$), and MAUVE. All metrics show strong Spearman rank correlations ($|\rho| \geq 0.80$)}
    \label{fig:classification_correlation}
\end{figure*}

\clearpage
\newpage
\section{Interpreting the Behavioral Clusters Results}
\label{appendix:interpreting_clusters}

\begin{table}[h]
  \caption{
    Top-50 distinctive n-grams for each behavioral category (Gemini 3.1 Pro, coding task). Each term is scored by its contrastive TF-IDF in the target category minus the average of the other two, and assigned exclusively to the category where it scores highest.}
  \label{tab:ngram_frequency}
  \centering
  \footnotesize
  \begin{tabular}{lp{\dimexpr\columnwidth-2cm}}
    \toprule
    \textbf{Category} & \textbf{Distinctive Terms} \\
    \midrule
    \textit{Well-Captured} & requesting alternative, recommendations, troubleshoot implementation, worked, problem current, rejects proposed, readability, alternative approaches, tool mixes, successful resolution, balances concise, moderate expansive, solution shares, anomalies, errors highly, proactively corrects, narrows solution, highly variable, logical flaws, length decreases, short fragmented, misunderstandings, expert, comparative, inconsistencies, implementation errors, state requesting, bug, starts polite, commands problem, proposes alternative, reports results, problem focused, complex logic, encountering errors, tests reports, solution reports, assistance modify, necessity, game, removal, actively tests, improvements, logical errors, mixes imperative, problem questions, refined iterative, decreases, logic errors, unprompted follow \\
    \midrule
    \textit{Missed} & does respond, analysis greetings, compressed raw, instruction followed, lack proactive, submission syntactic, did respond, message terse, usage imperative, respond, needs ends, analytical tool, progressed terse, conventions insufficient, highly exhaustive, authoritative, occurred terse, transactional analytical, lack correction, command fix, prompt did, respond evaluation, diagnostic tool, evaluation occurred, ended proactive, evaluate proactive, evolution greetings, mechanisms employed, instruction steering, response lack, determine evolution, transactional diagnostic, message greetings, transactional processing, progressed, needs prompt, reactive corrective, data determine, directed actions, actions direct, additional terse, actions steering, issues command, ends response, message progressed, determine change, needs does, steering mechanisms, occurred ended, evaluation occurs \\
    \midrule
    \textit{Hallucinated} & session direct, inquiry requesting, friendly inquisitive, add implementation, thanks awesome, consistent positivity, enthusiastic appreciative, monitors, actively monitors, positivity, hey expression, expresses agreement, thank select, enthusiastic satisfied, phrasing collaborative, friendliness, lot sense, makes lot, consistent friendliness, lot, followed feature, acknowledgment thanks, direct termination, thanks collaborative, builds partial, revelation terminated, enthusiastic curious, friendly engaged, current situation, phrasing social, immediately termination, hey thanks, actively processes, curious appreciative, enhance, dialogue moderate, evolution casual, thx, gratitude thx, consistent brief, agreement acknowledging, wow, directed expresses, termination structural, directed termination, thanks makes, thanks softeners, formal pleasantries, casual language, acknowledges solution \\
    \bottomrule
  \end{tabular}
\end{table}

\clearpage
\newpage
\section{Experiments Compute Resources}
\label{appendix:experiment_compute_resources}

All experiments were run on an internal cluster of NVIDIA H100 (80GB) and A40 (48GB) GPUs. For a single user simulator and task, conversation generation took approximately 3 hours on 8 H100s, or 4 H100s for closed-source simulators accessed via their APIs (the assistant runs on the H100s). User behavior description generation took approximately 2 hours on 4 H100s, and embedding took approximately 0.5 hours on 4 A40s. The full research project also included preliminary experiments that are not included in the final paper.

\section{License}
\label{appendix:license}

All code released with this project is under the Apache 2.0 License. The WildChat dataset is under the Open Data Commons Attribution License v1.0 (ODC-BY 1.0). All models are used under their respective licenses. Our use of existing artifacts is consistent with their intended use. The artifacts are all in English, and do not contain data with personally identifiable information.

\newpage
\section*{NeurIPS Paper Checklist}

\begin{enumerate}

\item {\bf Claims}
    \item[] Question: Do the main claims made in the abstract and introduction accurately reflect the paper's contributions and scope?
    \item[] Answer: \answerYes{} 
    \item[] Justification: The claims made in the abstract and introduction accurately reflect the paper’s contributions and scope.

\item {\bf Limitations}
    \item[] Question: Does the paper discuss the limitations of the work performed by the authors?
    \item[] Answer: \answerYes{} 
    \item[] Justification: Limitations are discussed in Appendix \ref{appendix:limitations}.

\item {\bf Theory assumptions and proofs}
    \item[] Question: For each theoretical result, does the paper provide the full set of assumptions and a complete (and correct) proof?
    \item[] Answer: \answerNA{} 
    \item[] Justification: There are no theoretical results in this paper.

    \item {\bf Experimental result reproducibility}
    \item[] Question: Does the paper fully disclose all the information needed to reproduce the main experimental results of the paper to the extent that it affects the main claims and/or conclusions of the paper (regardless of whether the code and data are provided or not)?
    \item[] Answer: \answerYes{} 
    \item[] Justification: The paper fully discloses all the information needed to reproduce the experimental results of the paper throughout Sections \ref{sec:method}, \ref{sec:experimental_setup}, \ref{sec:method_validation}, and \ref{sec:discussion}, and Appendix \ref{appendix:user_behavior_representations}, \ref{appendix:user_goal_classification}, and \ref{appendix:conversation_generation_prompts}. Additionally, our code has been released to ensure our experiments are easily reproducible.

\item {\bf Open access to data and code}
    \item[] Question: Does the paper provide open access to the data and code, with sufficient instructions to faithfully reproduce the main experimental results, as described in supplemental material?
    \item[] Answer: \answerYes{} 
    \item[] Justification: The paper provides access to the data and code, and includes sufficient instructions for how to faithfully reproduce the main experimental results throughout Sections \ref{sec:method}, \ref{sec:experimental_setup}, \ref{sec:method_validation}, and \ref{sec:discussion}, and Appendix \ref{appendix:user_behavior_representations}, \ref{appendix:user_goal_classification}, and \ref{appendix:conversation_generation_prompts}.

\item {\bf Experimental setting/details}
    \item[] Question: Does the paper specify all the training and test details (e.g., data splits, hyperparameters, how they were chosen, type of optimizer) necessary to understand the results?
    \item[] Answer: \answerYes{} 
    \item[] Justification: The paper specifies all the experimental details needed to reproduce the experimental results of the paper throughout Sections \ref{sec:method}, \ref{sec:experimental_setup}, \ref{sec:method_validation}, and \ref{sec:discussion}, and Appendix \ref{appendix:user_behavior_representations}, \ref{appendix:user_goal_classification}, and \ref{appendix:conversation_generation_prompts}.

\item {\bf Experiment statistical significance}
    \item[] Question: Does the paper report error bars suitably and correctly defined or other appropriate information about the statistical significance of the experiments?
    \item[] Answer: \answerYes{} 
    \item[] Justification: We report Spearman rank correlations in our ablation study in Section \ref{sec:method_validation_ablations}, and Appendix \ref{appendix:ablation_results} and \ref{appendix:linear_classification}.

\item {\bf Experiments compute resources}
    \item[] Question: For each experiment, does the paper provide sufficient information on the computer resources (type of compute workers, memory, time of execution) needed to reproduce the experiments?
    \item[] Answer: \answerYes{} 
    \item[] Justification: Sufficient information on compute resources are provided in Appendix \ref{appendix:experiment_compute_resources}
    
\item {\bf Code of ethics}
    \item[] Question: Does the research conducted in the paper conform, in every respect, with the NeurIPS Code of Ethics \url{https://neurips.cc/public/EthicsGuidelines}?
    \item[] Answer: \answerYes{} 
    \item[] Justification: The research in the paper conform, in every respect, with the NeurIPS Code of Ethics.

\item {\bf Broader impacts}
    \item[] Question: Does the paper discuss both potential positive societal impacts and negative societal impacts of the work performed?
    \item[] Answer: \answerYes{} 
    \item[] Justification: Yes, the paper addresses the broader impacts of the work.
    
\item {\bf Safeguards}
    \item[] Question: Does the paper describe safeguards that have been put in place for responsible release of data or models that have a high risk for misuse (e.g., pre-trained language models, image generators, or scraped datasets)?
    \item[] Answer: \answerNA{} 
    \item[] Justification: The paper poses no such risks.

\item {\bf Licenses for existing assets}
    \item[] Question: Are the creators or original owners of assets (e.g., code, data, models), used in the paper, properly credited and are the license and terms of use explicitly mentioned and properly respected?
    \item[] Answer: \answerYes{} 
    \item[] Justification: The creators and original owners of assets used in the paper are properly credited and the terms of use are explicitly mentioned in Appendix \ref{appendix:license} and properly respected.

\item {\bf New assets}
    \item[] Question: Are new assets introduced in the paper well documented and is the documentation provided alongside the assets?
    \item[] Answer: \answerYes{} 
    \item[] Justification: We document all released assets, including code and prompts.

\item {\bf Crowdsourcing and research with human subjects}
    \item[] Question: For crowdsourcing experiments and research with human subjects, does the paper include the full text of instructions given to participants and screenshots, if applicable, as well as details about compensation (if any)? 
    \item[] Answer: \answerYes{}{} 
    \item[] Justification: Details about our human study are provided in Section \ref{sec:method_validation_human_study} and \ref{appendix:humman_study}.

\item {\bf Institutional review board (IRB) approvals or equivalent for research with human subjects}
    \item[] Question: Does the paper describe potential risks incurred by study participants, whether such risks were disclosed to the subjects, and whether Institutional Review Board (IRB) approvals (or an equivalent approval/review based on the requirements of your country or institution) were obtained?
    \item[] Answer: \answerYes{} 
    \item[] Justification: In Section \ref{sec:method_validation_human_study}, we describe how our human study was declared exempt by our Institutional Review Board (IRB).

\item {\bf Declaration of LLM usage}
    \item[] Question: Does the paper describe the usage of LLMs if it is an important, original, or non-standard component of the core methods in this research? Note that if the LLM is used only for writing, editing, or formatting purposes and does \emph{not} impact the core methodology, scientific rigor, or originality of the research, declaration is not required.
    \item[] Answer: \answerYes{} 
    \item[] Justification: In Section \ref{sec:discussion_interpretation}, we describe that LLMs were used to identify meaningless terms to reduce noise during the TF-IDF analysis.

\end{enumerate}
\newpage

\end{document}